\documentclass[lettersize,journal]{IEEEtran}
\usepackage{amsmath,amsfonts}
\usepackage{algorithm}
\usepackage{array}
\usepackage{textcomp}
\usepackage{stfloats}
\usepackage{url}
\usepackage{verbatim}
\usepackage{graphicx}
\usepackage{cite}
\usepackage{color}
\usepackage{xcolor}
\usepackage{subfigure}
\usepackage{graphics}
\usepackage{multirow}
\newcommand{\eg}{\textit{e}.\textit{g}. }

\usepackage{times}
\usepackage{epsfig}
\usepackage{graphicx}
\usepackage{amsmath}
\usepackage{amssymb}
\usepackage{color}
\usepackage{algorithm}
\usepackage{algpseudocode}
\usepackage{dsfont}
\usepackage{cite}
\usepackage{booktabs}
\usepackage{tabularx}
\usepackage{url}
\usepackage{soul}
\usepackage{balance}
\usepackage{paralist} % compactitem

\usepackage{lscape}
\usepackage{array,booktabs}

\usepackage{rotating}

\usepackage[justification=centering]{caption}
\begin{document}

\title{RGBD Object Tracking: An In-depth Review}

\author{Jinyu Yang$^{1}$,~\IEEEmembership{Student Member,~IEEE},
        Zhe Li$^{1}$,~\IEEEmembership{Student Member,~IEEE},
        Song Yan,~\IEEEmembership{Student Member,~IEEE},\\
        Feng Zheng$^{*}$,~\IEEEmembership{Member,~IEEE},
        Ale\v{s} Leonardis,~\IEEEmembership{Member,~IEEE},
        Joni-Kristian K\"am\"ar\"ainen,~\IEEEmembership{Member,~IEEE},\\
        Ling Shao,~\IEEEmembership{Fellow,~IEEE}

%\thanks{Manuscript received XXX XX, 20XX. (Corresponding author: Feng Zheng).}
\thanks{J. Yang is with  Southern University of Science and Technology, Shenzhen 518055, China, and also with University of Birmingham, Birmingham B15 2TT, U.K. (e-mail: jinyu.yang96@outlook.com).}
\thanks{Z. Li and F. Zheng are with Southern University of Science and Technology, Shenzhen 518055, China (e-mail: liz8@mail.sustech.edu.cn and f.zheng@ieee.org).}
\thanks{S. Yan and J.K. K\"am\"ar\"ainen are with Tampere University, Tampere 33720, Finland. (e-mail: \{song.yan, joni.kamarainen\}@tuni.fi).}
\thanks{A. Leonardis is with the University of Birmingham, Birmingham B15 2TT, U.K. (e-mail: a.leonardis@cs.bham.ac.uk).}
\thanks{L. Shao is with the Inception Institute of Artificial Intelligence, Abu Dhabi, United Arab Emirates. (e-mail: ling.shao@ieee.org)}
\thanks{$^1$The first two authors contributed equally. $^*$Corresponding author.}
}

        % <-this % stops a space
%\thanks{This paper was produced by the IEEE Publication Technology Group. They are in Piscataway, NJ.}% <-this % stops a space
%\thanks{Manuscript received April 19, 2021; revised August 16, 2021.}
    % \thanks{Jinbao Wang and Feng Zheng are with the Department of Computer Science and Engineering, Southern University of Science and Technology, Shenzhen 518055, China (e-mail: linkingring@163.com;  f.zheng@ieee.org).}
    % \thanks{L. Shao is with the Inception Institute of Artificial Intelligence, Abu Dhabi, The United Arab Emirates. (e-mail: ling.shao@ieee.org)}

\markboth{Journal of \LaTeX\ Class Files,~Vol.~14, No.~8, August~2021}%
{Shell \MakeLowercase{\textit{et al.}}: A Sample Article Using IEEEtran.cls for IEEE Journals}


\maketitle
%no thorough review has been provided to summarize this area for the computer vision community.
%Particularly, we provide in-depth analysis on depth information quality and depth favorable scenarios for the first time.
%against object loss.
\begin{abstract}
RGBD object tracking is gaining momentum in computer vision research thanks to the development of depth sensors.
Although numerous RGBD trackers have been proposed with promising performance, an in-depth review for comprehensive understanding of this area is lacking.
In this paper, we firstly review RGBD object trackers from different perspectives, including RGBD fusion, depth usage, and tracking framework.
Then, we summarize the existing datasets and the evaluation metrics.
We benchmark a representative set of RGBD trackers, and give detailed analyses based on their performances.
Particularly, we are the first to provide depth quality evaluation and analysis of tracking results in depth-friendly scenarios in RGBD tracking.
For long-term settings in most RGBD tracking videos, we give an analysis of trackers' performance on handling target disappearance.
To enable better understanding of RGBD trackers, we propose robustness evaluation against input perturbations.
Finally, we summarize the challenges and provide open directions for this community.
All resources are publicly available at \url{https://github.com/memoryunreal/RGBD-tracking-review}.

\end{abstract}


\section{Introduction}
%task definition
% \iffalse
% Visual object tracking (VOT) aims at localizing an object in a video sequence given the object description (center location and scale) in the first frame.
Visual object tracking (VOT) aims at localizing an arbitrary object in a video sequence given the object description (center location and scale) in the first frame.
% \lizhe{Visual object tracking (VOT) aims at obtaining the state of an object in the subsequent video frames by giving the initial state (center location and scale) of the object in the first frame.}
%
%
% As a class-agnostic task, VOT focuses on tracking an arbitrary object with only the initial description in the first frame.
%
%
% \lizhe{As a class-agnostic task, VOT focuses on tracking an arbitrary object with only the initial state described in the first frame.}
%applications
Recent years have witnessed a great development of object tracking due to its diverse applications, \textit{e.g.}, autonomous driving \cite{luo2018fast}, and robotics \cite{machida2012human}.
%and visual surveillance \cite{hampapur2005smart}.
% \iffalse Problems like occlusion, deformation, rotation remain this field challenging, which have been widely investigated by researchers. \fi
Problems like object occlusion, deformation, rotation, are still challenging in this field and have been extensively investigated by researchers.
%development
A large amount of RGB trackers, especially short-term trackers, emerge to boost this community.
Since 2013, Kernelized Correlation Filter (KCF) \cite{2015High} was introduced to solve the template matching problem.
% \lizhe{Since 2013, the advent of Kernelized Correlation Filters (KCF) introduced solving the template matching problem into the visual object tracking field.}
Due to the popularity of deep neural networks in computer vision, Siamese network and deep correlation filter based trackers are very popular in object tracking \cite{2016Fully,2019ATOM,zhang2019deeper,bhat2019learning,zhu2018distractor,2018High,li2019siamrpn++}.
Progress in RGB tracking has been further boosted by the emergence of standard datasets and evaluation protocols.
There are diverse and large datasets for model training and evaluation, such as OTB \cite{OTBPAMI2015}, GOT-10k \cite{Lianghua2019GOT} and TrackingNet \cite{2018TrackingNet}.
This field is also fueled by the annual VOT challenges \cite{2016The, 2015The, 2017The, kristan2019seventh, Kristan2020a,kristan2021ninth}.

\begin{figure*}
	\centering
		\includegraphics[width=\linewidth]{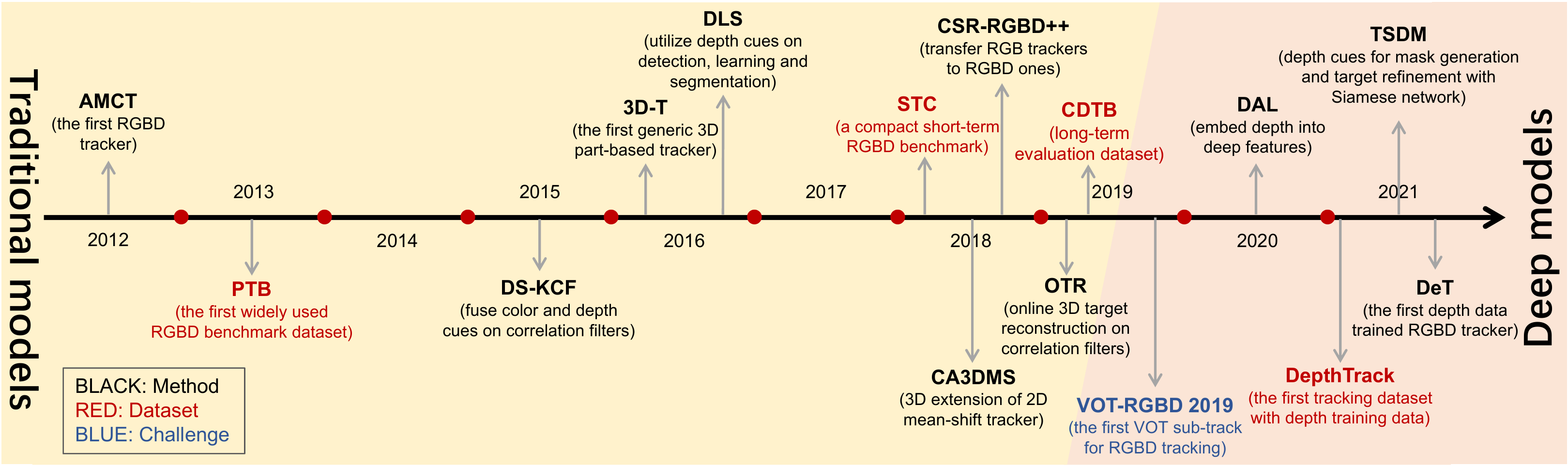}
	\caption{Chronology of RGBD object tracking.}
	\label{fig_chro}
\end{figure*}

%extend to rgbd tracking
Although significant progress has been made in RGB-based tracking, there are still tracking failures that are hard to be solved by color information. %intractable with color information only.
Therefore, other modalities are added to provide complementary information, including depth, thermal, and event information \cite{yan2021depthtrack,2021LasHeR,DBLP:journals/tmm/XuMLL22,wang2021viseventbenchmark}.
Among them, RGBD (RGB + Depth) object tracking is gaining momentum in the past decade thanks to the affordable advanced depth cameras, such as Microsoft Kinect and Intel RealSense.
%use of depth
On the one hand, depth maps provide essential cues on occlusion reasoning and depth-based object segmentation\cite{DBLP:journals/tmm/XieRGWWW18,DBLP:journals/tmm/LiWZYL19}.
For example, CA3DMS \cite{liu2018context} uses a context-aware 3D mean-shift to handle occlusion, and DM-DCF \cite{kart2018depth} proposes a depth-based segmentation to train a constrained Discriminative Correlation Filter (DCF).
On the other hand, RGBD channels can sense both appearance and geometric components for better object-and-background separation.
For example, OTR \cite{Kart_CVPR_2019} uses both color and depth information to build a spatial reliability map and reconstruct an object 3D model.
Therefore, exploring the depth cue is indeed helpful for multi-modal tracking.
%development
Early RGBD trackers utilize direct heuristic extensions of RGB-based methods, which tend to extract hand-crafted features from depth maps to solve specific challenges.
% \lizhe{Early RGBD trackers exploited some intuitive and heuristic extensions to RGB-based methods that tend to extract hand-crafted features from depth images to address specific challenges.}
For example,
% \iffalse PT\cite{song2013tracking} gives a set of baseline RGBD trackers including HOG, optical flow and point clouds. \fi
% \lizhe{
PT \cite{song2013tracking} introduces a set of RGBD baseline trackers, including a traditional 2D tracker with additional depth HOG features, a 2D optical flow tracker, and a 3D point cloud tracker.
Recently, deep networks are also introduced to RGBD tracking, but they are still straightforward extensions of RGB baselines \cite{zhao2021tsdm}\cite{yan2021depthtrack}.
At the same time, the annual VOT challenge\cite{kristan2019seventh} has had a specific track for RGBD input since 2019.
By providing a test set and evaluation protocols, RGBD tracking can gain more attention in the object tracking community.
% \cite{kristan2019seventh,Kristan2020a,kristan2021ninth}.
Until now, there have been 13 participant RGBD trackers in the VOT-RGBD challenges\cite{kristan2019seventh,Kristan2020a,kristan2021ninth}.
According to VOT reports, the VOT trackers show superior performance over traditional solutions, and continuously improve the state-of-the-arts.
A brief chronology of RGBD tracking is shown in Fig.~\ref{fig_chro}.

%However, compared to RGB-based object tracking which has been a mature and booming community, the development of RGBD object tracking still lags behind a lot.
%\subsection{Scope}
As a newly developing area, research on RGBD tracking remains independent and non-systematic.
There are many interesting topics required to be investigated as follows.
%\begin{enumerate}
%\itemsep=0pt
\begin{itemize}
\item How does depth information benefit tracking?
According to conclusions from VOT-RGBD challenges, whether depth helps to track and what kind of methods are beneficial remains as an open question.

\item How does depth data quality affect the tracking performance?
The quality of depth maps is a vital issue, but its impact on tracking has not been investigated before.

\item What kinds of scenarios need help from depth information?
Finding out the depth favorable scenarios can be meaningful for a wide range of applications in RGBD tracking and other related tasks.

\item How does the long-term setting reflect on RGBD tracking?
As RGBD tracking tends to be a long-term task, the effect of the target disappearance and reappearance requires to be investigated.

\item How about the robustness of current tracking models?
For RGBD tracking models, with inputting RGB and depth channels separately, a tiny perturbation on input can be of vital importance to tracking performance.
However, how trackers perform with different input perturbations is an unexplored area.
\end{itemize}
%\end{enumerate}

This survey aims to provide a thorough review of this field and give in-depth analysis from various views.
Our empirical analysis will answer the above questions and
% aim to exploit the opportunity of
exploit the potential opportunity of
using the depth components in tracking.
%In addition,
Specifically, we restrict this survey to RGBD single object tracking.
Some research areas are related but not covered here including multi-object tracking, RGB + Thermal tracking, 3D tracking as well as scene understanding \cite{DBLP:journals/tmm/GaoPWG22,DBLP:journals/tmm/LiWXQ21}.
% Although there are multiple surveys on object tracking, in this paper, we focus on reviewing the existing RGBD based tracking models and benchmark datasets.
% \lizhe{Redundancy?  Specifically, this survey mainly focuses on the major progress in this area and gives in-depth analysis on various aspects.}
%For completeness and better readability, some related works are also included.

\begin{table*}
\caption{Summary of existing surveys in related fields.}
\setlength\tabcolsep{2.6pt}
\begin{tabular*}{\linewidth}{|l|l|l|l|}
\hline%
\multicolumn{1}{|c|}{Title}& \multicolumn{1}{c|}{Scope} &Year&\multicolumn{1}{c|}{Venue}\\%
\hline%
%Multimodal fusion for multimedia analysis: a survey\cite{2010Multimodal}&&2010\\
Recent Advances on Multicue Object Tracking: A Survey \cite{2016Recent} &Multimodal object tracking &2016&AI Review\\
Multiple Human Tracking in RGB-D Data: A Survey \cite{2016Multiple}&RGBD multi-human tracking&2016&IET Computer Vision\\
RGBD Datasets: Past, Present, and Future \cite{2016RGBD} &RGBD dataset &2016 &CVPR Workshop\\
RGB-D Datasets Using Microsoft Kinect or Similar Sensors: A Survey \cite{2017RGB}&RGBD dataset &2017&MTA\\%
%Multimodal Machine Learning: A Survey and Taxonomy\cite{2017Multimodal}&Multimodal machine learning&2017&TPAMI\\
Object Fusion Tracking Based on Visible and Infrared Images: A Comprehensive Review \cite{2020Object}&RGBT object tracking&2020&Information Fusion\\
Multi-modal Visual Tracking: Review and Experimental Comparison \cite{2020Multi}&Multimodal object tracking&2020&Submitted to PR\\
\hline
\end{tabular*}
\label{survey}
\end{table*}

\subsection{Related reviews and surveys}
Table \ref{survey} lists several previous related reviews and surveys.
Walia \textit{et al.} \cite{2016Recent} reviewed single-modal and multi-modal tracking methods preceding 2016.
Camplani \textit{et al.} \cite{2016Multiple} reviewed methods for tracking multiple humans with RGBD data.
In 2016 and 2017, there were two surveys \cite{2016RGBD,2017RGB} on existing RGBD datasets used in different applications, including object recognition, semantic reasoning and segmentation, human recognition, and pose estimation.
%, \textit{etc}.
Zhang \textit{et al.} \cite{2020Object} focused on RGB and infrared tracking and outlined recent progress on RGBT tracking methods.
Finally, a more recently published survey \cite{2020Multi} covers both RGBT and RGBD tracking works until 2020. %, and provided a more substantial discussion in a large scope.

Different from the above surveys, which focus on earlier datasets and algorithms, or related areas, this work systematically and comprehensively reviews RGBD object tracking from multiple perspectives.
%To the best of our knowledge, this is the first systematic review in this field.
The rest of this paper is organized as follows.
Sec.~\ref{model} summarizes existing RGBD tracking algorithms.
Datasets and evaluation metrics are reviewed in Sec.~\ref{dataset_metric}.
Then, we conduct extensive experiments in Sec.~\ref{experiments}, in which we compare overall performance (Sec.~\ref{compare}), examine the effect of depth data quality (Sec.~\ref{quality}), find out depth-related scenarios (Sec.~\ref{favorable}), and give analysis on long-term setting (Sec.~\ref{longterm}).
Moreover, we evaluate robustness for RGBD trackers against input perturbations (Sec.~\ref{robustness}).
Finally, Sec.~\ref{discussion} gives in-depth discussions on future directions.

\subsection{Contributions}
Our contributions are:
\begin{itemize}
\item \textbf{A systematic review of RGBD tracking models, datasets, and evaluation metrics.}
We categorize existing RGBD trackers from different perspectives, including fusion strategy, depth usage, and feature extraction.
All existing RGBD object tracking benchmarks and evaluation metrics are reviewed as well.
\item \textbf{Performance evaluation of RGBD trackers.}
We compile a hybrid dataset and provide extensive experiments on 18 representative trackers, based on which we give detailed analysis to unfold the pros and cons of current RGBD datasets and trackers.
%We reproduce 18 advanced RGBD tracking methods and have some interesting findings. %on the variation in tracker performance on videos with object loss properties.
\item \textbf{An in-depth experimental analysis for RGBD object tracking from various aspects.}
We investigate several important components, including the effect of depth data quality, attribute-based analysis, and long-term setting effects.
In particular, we are the first to study robustness to input perturbations in RGBD object tracking.
\item \textbf{Overview of challenges and open directions.}
From our empirical analysis of RGBD object tracking, we thoroughly investigate the challenges for RGBD tracking and provide potential directions for future research.
\end{itemize}

\section{Models and Taxonomy}\label{model}
%All of the RGBD tracking baselines are representative RGB tracking paradigms.
To systematically review RGBD trackers, we categorize the models.
There are mainly three taxonomies: RGB and depth fusion paradigm, depth usage, and feature extraction method.
We specifically focus on the trackers participants in VOT RGBD tracking challenges.
In the following, several representative models in each taxonomy will be described. %illustrated and exemplified in different subsections.
%Specifically, we give taxonomy from whether the tracker can be embedded; how RGB and depth channels fuse; how trackers extract features from depth; 2D or 3D
Table~\ref{tbl_model_statistics} summarizes existing RGBD trackers.
%\subsection{Embedding/non-embedding depth module}
%We here review some representative works.

\begin{table*}%[width=.9\linewidth,cols=6,pos=h]
\begin{center}
\caption{Statistics of RGBD tracking models.
``CL/DL'' indicates whether the tracker is a classical/deep-learning based method.
``ST/LT'' indicates short-term/long-term trackers.
``Occ.Han.'' indicates occlusion handling.}
\label{tbl_model_statistics}
\setlength\tabcolsep{6pt}
\begin{tabular*}{\linewidth}{| c|c|c|c|c|c|c|c|c|c |}
\hline
%Category
 Method & Year & Publication & CL/DL & Framework &Backbone & Training data & ST/LT &Occ.Han. &Code \\%& Depth Purpose\\
\hline
%\multirow{22}{Published Trackers}
AMCT \cite{GM2012Adaptive} &2012 & JDOS &CL &Condensation &-&- &ST& & \\
 \hline
 PT\cite{song2013tracking} &2013& CVPR & CL& SVM &- &- & ST &&\\ % & Occlusion detection \\
 \hline
 MCBT \cite{2014Multi} &2014 & Neurocomputing &CL & &-&-&LT&&\\
 \hline
 DS-KCF\cite{camplani2015real} &2015 & BMVC & CL& KCF &-  &- & ST & & \checkmark\\% & Target segmentation; Occlusion detection \\
 OL3DC \cite{2015Online} &2015 & Neurocomputing &CL &SURF &-&-&LT &\checkmark &\\
 CDG \cite{2016Using} &2015 & CAC &CL &SVM &- &- &LT &\checkmark& \\
 DOHR \cite{2016Robust} &2015 &FSKD &CL&Bayesian &- &-&LT & \checkmark &\\
 ISOD \cite{20153D} &2015 & Signal Processing &CL& &- &- &LT & \checkmark &\\
 \hline
 DS-KCF\_shape\cite{2016DS} & 2016&J.RTIP & CL& KCF & -&- & ST & & \checkmark \\%& Target segmentation; Occlusion detection \\
 3D-T \cite{Bibi_2016_CVPR} &2016 & CVPR & CL& KCF &- &- & LT &\checkmark &\\%& Occlusion handling\\
 OAPF\cite{MESHGI201681} &2016 & CVIU & CL& Particle Filter& - &-& LT &\checkmark &\\%& Occlusion detection\\
 DLS \cite{an2016online}  &2016 &ICPR&CL&KCF&- &- & LT & \checkmark &\\%&Target segmentation\\
 \hline
 ODIOT\cite{zheng2017online} &2017 & Neural Processing Letters & CL & TLD &- &- & LT &\checkmark & \\%& Occlusion handling\\
 ROTSL \cite{2017Robust} &2017 &ITEE &CL&Particle filter &-&- &LT &\checkmark& \\
 \hline
 STC\cite{xiao2017robust} &2018&IEEE TCYB & CL& KCF& - &- & ST & & \checkmark \\%&  BBox position fusion\\
 CSR\_RGBD++ \cite{Kart_ECCVW_2018}&2018 & ECCVW & CL& CSR\_DCF &-  & - & LT &\checkmark & \checkmark \\%& Foreground segmentation\\
 DM-DCF\cite{kart2018depth} &2018 & ICPR & CL & CSR\_DCF&- &- & ST & & \\% & Mask generation\\
 SEOH \cite{2018Real} &2018 & IEEE Access &CL &KCF&-&- &LT &\checkmark &\\
 OACPF \cite{8463446} &2018 & IEEE Access &CL &Particle filter&-&- &LT &\checkmark &\\
 RT-KCF \cite{0A} &2018 &CCDC &CL &KCF &-&-&LT &\checkmark & \\
 CA3DMS \cite{liu2018context} &2018 &IEEE TMM & CL &  MeanShift&-  &- & LT &\checkmark &\checkmark  \\%& Point clouds\\
 \hline
 OTR \cite{Kart_CVPR_2019} &2019 & CVPR & CL & CSR\_DCF &- &- & LT &\checkmark & \checkmark \\
 Depth-CCF\cite{li2019depth} &2019& IOP & CL &DCF &- &- & LT &\checkmark & \\
 ECO\_TA \cite{8752050} &2019 &IEEE Sensors &CL &ECO &- &- & ST &\checkmark &\\
 RGBD-OD\cite{xie2019rgb} &2019& CIS & CL &  & PointNet&Point pre-trained & LT &\checkmark & \\
 H-FCN\cite{jiang2019hierarchical} &2019& Information Fusion & CL&ECO &-  &- & LT &\checkmark & \\
 3DMS \cite{8861628} &2019& ICST & CL &Mean Shift &- &- &LT &\checkmark & \\
 \hline
 WCO\cite{8950173} &2020 &IEEE Sensors &CL&DCF &- &- &ST & &\\
 RF-CFF\cite{wang2020robust} &2020&Applied Soft Computing & DL &KCF & VGG-Net & RGB pre-trained & LT &\checkmark &\\% & BBox position fusion  \\
 SiamOC\cite{zhang2020occlusion} & 2020 &ICSP & DL & SiamDW-RPN &ResNet-18 & RGB pre-trained & LT &\checkmark & \\%& Occlusion detection \\
 DAL\cite{2019DAL} &2020 &ICPR & DL & ATOM &ResNet-18 &RGB pre-trained & ST& & \checkmark \\% &  Foreground segmentation; Feature fusion\\
 \hline
 3s-RGBD \cite{xiao2021single}& 2021&Neurocomputing & DL & SiamFC &AlexNet &RGB pre-trained &LT &\checkmark &~ \\%& ST &Occlusion handling  \\
 TSDM\cite{zhao2021tsdm} &2021 &ICPR & DL & SiamRPN++ &ResNet-50 &RGB pre-trained & ST & & \checkmark \\%& Mask generation; BBox refinement\\
 DeT\cite{yan2021depthtrack} &2021 & ICCV & DL& DiMP& ResNet-50 &RGB + RGBD &ST & & \checkmark \\%& feature fusion\\
\hline
 ATCAIS19 &2019& VOT-2019 & DL & ATOM &ResNet-18 &RGB pre-trained & LT& \checkmark & \checkmark  \\%& Occlusion detection and target redetection\\
 SiamDW\_D &2019& VOT-2019 & DL & SiamDW &ResNet-50 &RGB pre-trained & LT & \checkmark & \checkmark \\%& Disappearance estimation\\
 SiamM\_Ds &2019& VOT-2019 & DL & SiamMask &ResNet-50 &RGB pre-trained & LT &\checkmark  & \checkmark \\%& Target location constraint\\
 LTDSEd &2019& VOT-2019 & DL & LT-DSE &ResNet-50 &RGB pre-trained & LT &\checkmark & \checkmark \\%& -\\
 ATCAIS20 &2020& VOT-2020 & DL & ATOM &ResNet-18 &RGB pre-trained & LT &\checkmark & \checkmark \\%& Occlusion detection\\
 CLGS\_D &2020& VOT-2020 & DL & SiamMask &ResNet-50 &RGB pre-trained & LT &\checkmark & \checkmark \\%& BBox refinement\\
 DDiMP &2020& VOT-2020 & DL & SuperDiMP &ResNet-50 &RGB pre-trained & ST & & \checkmark \\%& Prevent scale change \\
 Siam\_LTD &2020& VOT-2020 & DL& SiamRPN &ResNet-50  &RGB pre-trained & LT & \checkmark & \checkmark \\%& -\\
 sttc\_rgbd &2021 & VOT-2021 & DL & STARK &ResNet-50 &RGB pre-trained & ST & \checkmark & \checkmark \\%&  -\\
 STARK\_RGBD &2021 & VOT-2021 & DL & STARK &ResNet-50 &RGB pre-trained & LT &\checkmark & \checkmark \\%& - \\
 SLMD &2021& VOT-2021 & DL & PrDiMP &ResNet-50 &RGB pre-trained & LT & & \checkmark \\%& - \\
 TALGD &2021& VOT-2021 & DL & SuperDiMP &ResNet-50 &RGB pre-trained & LT & \checkmark& \checkmark \\%& Occlusion or disappearance reasoning; target retrieval\\
 DRefine &2021& VOT-2021 & DL & SuperDiMP &ResNet-50 &RGB pre-trained & ST & & \checkmark \\%& Consistency judgement\\
\hline
\end{tabular*}
\end{center}
\end{table*}

\subsection{Early/late RGBD fusion}
Existing fusion strategies in RGBD tracking can be divided into early fusion and late fusion.

\subsubsection{Early fusion}
Early fusion based models generally follow two strategies:
RGB and depth images are first fed into independent feature extraction modules separately and then combined. %, and their feature representations are combined as a joint representation.
Then the combined feature map is used to obtain a final prediction, such as DS-KCF \cite{camplani2015real}, CSR\_RGBD++ \cite{kart2018depth}, ECO\_TA \cite{8752050}.
Another strategy is to integrate the depth channel with RGB channels to form a four-channel input, such as PT~\cite{song2013tracking}.
For early fusion, MCBT \cite{2014Multi} combines optical flow, color, and depth information, which are simultaneously incorporated to predict the precise position.
Camplani \textit{et al.} proposed DS-KCF \cite{camplani2015real} which utilized HOG features for both color and depth maps.
OAPF \cite{MESHGI201681} employs multiple features, \textit{e.g.} HoG, from color and depth streams to improve robustness against illumination changes and clutter and boost the performance.
%Another way is to fuse the high-level feature from independent streams with independent networks,
Another example is DeT~\cite{yan2021depthtrack}, which extracts depth features by using an additional depth branch.
With fusing the color feature and depth feature, it can use ATOM \cite{2019ATOM} or DiMP \cite{bhat2019learning} tail part to perform the actual tracking.

%CSR\_RGBD++\cite{kart2018depth}, ECO\_TA\cite{8752050}.

\subsubsection{Late fusion}
Late fusion based models generally process both modalities simultaneously, and the independent models for RGB stream and depth stream are built to make decisions.
For example, Xiao \textit{et al.} proposed STC~\cite{xiao2017robust} which fuses two single-modal trackers through weighted maps.
CDG \cite{2016Using} uses depth gradient information to extract depth motion models and weights the results from RGB and depth models.
RT-KCF \cite{0A} fuses response maps instead of features to get a good performance.
RF-CFF \cite{wang2020robust} focuses on fusing tracking results of RGB and depth images, in which objects are tracked in RGB and depth images separately using the correlation filter.
% Then, the results in RGB and depth images are adaptively fused.
Then results from RGB and depth images are adaptively fused.
%the results of hierarchical CNN features are adaptively predicted according to comparison of results of forward and backward tracking

\subsection{2D/3D depth usage}
%treat depth cues as 2D structure or 3D structure.
%According to the ways how researchers treat depth cue
Since depth maps provide appearance descriptions and geometry information for tracked objects, there are some methods treating depth maps in 2D and 3D structures, respectively.

\subsubsection{2D usage}
% In 2D view,
Depth map naturally provides a texture-free segmentation between foreground and background, so it is common to use depth cues
% histogram
for object segmentation.
%%%% Song: DeT may be not suitable for segmentation ?? %%%
% DeT \cite{yan2021depthtrack} utilizes colormaps of depth maps to transfer the depth information into colored ones.
%
%Representative models include DeT \cite{yan2021depthtrack} and DM-DCF \cite{kart2018depth}.
In CSR\_RGBD++ \cite{Kart_ECCVW_2018}, a depth augmented foreground segmentation is formulated by graph cut to obtain a foreground mask in the target update.
% In DM-DCF \cite{kart2018depth}, the depth-based segmentation masks are extracted to train a constrained DCF.
DM-DCF \cite{kart2018depth} extracts the depth-based segmentation masks in order to train a constrained DCF.

\subsubsection{3D usage}
% In 3D space,
Depth information provides the 3D spatial description of objects.
Representative trackers include OTR \cite{Kart_CVPR_2019}, CA3DMS \cite{liu2018context} and 3D-T \cite{Bibi_2016_CVPR}.
Among them, 3D-T \cite{Bibi_2016_CVPR} is the first 3D part-based tracker, which exploits parts to preserve temporal structural information and helps in particle pruning.
CA3DMS \cite{liu2018context} is a 3D extension of the classical mean-shift tracker, aiming to address two shortboards of 3D mean-shift: the online adaption and total occlusion handling.
OTR \cite{Kart_CVPR_2019} implements online 3D object construction to learn a robust view-specific discriminative correlation filter (DCF), extending the 2D tracking structure to 3D representation.
The 3D construction
% is to enhance
benefits the tracking performance from two aspects: generation of spatial description for the constrained 2D DCF learning; 3D pose estimation based on point clouds to localize the object after heavy occlusion.
%Before VOT-RGBD challenges, OTR leads the leaderboard of the RGBD benchmarks.

\subsubsection{Mixed usage}
There are also hybrid trackers which jointly use 2D and 3D models to combine 2D appearance features and 3D spatial features.
For example, DLS \cite{an2016online} simultaneously builds two target models: a 2D appearance model built upon
% the commonly used image
the features extracted from both color and depth frames, and a 3D distribution model built according to the point cloud distribution on the target surface.
The depth histogram is used to adaptively segment the target in depth frames and project cloud points into a depth image patch.
SEOH \cite{2018Real} employs the spatial continuity in depth values for scale estimation, and a part-based model updating strategy to deal with occlusion.
Another representative is TSDM \cite{zhao2021tsdm},
% TSDM consists
consisting of an RGB tracking core and two assistant modules.
First, the core is SiamRPN++~\cite{li2019siamrpn++}, which takes an image pair (template and mask images) as input.
Then, a mask-generator module utilizes depth information to generate a mask image for a candidate search image.
%, which can reduce the interference of background distractors and clear out some image information irrelevant to the target.
%
% Finally, a depth-refiner module improves the tracker performance by cutting out non-target areas from the original outputs and gives a smaller and more precise mask.
%
Finally, a depth-refiner module cuts out non-target areas from the original outputs and gives a smaller and more precise mask.

\subsection{Heuristic/deep models}
% Useful cues provided by depth information, like boundary cues, can help identify object characteristics.
Depth information provides useful cues, \eg boundary cues, and helps to identify object characteristics.
Over the past several years, many traditional
% tracking models
trackers with handcrafted features are designed by using these specific cues. % uses.
%as an early attempt, the gray-scale scalable gradient features of can be applied to the depth layer.
For example,
MCBT \cite{2014Multi} uses the depth mean and variance in the target region to measure the difference between candidates and templates.
PT \cite{song2013tracking} proposes a series of baseline trackers with handcrafted features, including a traditional 2D image patch-based tracker, a 3D point cloud-based tracker, and a low-level optical flow-based tracker.
STC \cite{xiao2017robust} first uses depth HOG as depth features, and then the RGB and depth features are separately used in KCF to find the target position in a global layer.
% Specifically,
As shown in Table~\ref{tbl_model_statistics}, there are many methods specially designed for occlusion handling since depth cues straightforward indicate the target locations.
ISOD \cite{20153D} exploits depth information obtained from binocular video data to detect occlusion, which prevents improper appearance model updating during occlusions.
Meshgi \textit{et al.}~\cite{MESHGI201681} proposed an occlusion aware particle filter framework that employs a probabilistic model with a latent variable to represent an occlusion flag.
Liu \textit{et al.}~\cite{liu2018context} proposed a context-aware 3D Meanshift method, which compares
% two
depth differences between the target
% before disappearing
and the occluder, and between the nearby 3D point and the occluder to detect and recover from tracking failures caused by full occlusions.
CSR\_RGBD++ \cite{Kart_ECCVW_2018} sets multiple assumptions in the occlusion recovery stage (the positions are similar before disappearing and after disappearing,
the target speed remains constant)
% the speed of the object remains constant) %, etc.)
to recapture the object.
% \noindent
% \textbf{SiamOC \cite{zhang2020occlusion}.} An occlusion estimation module and target location correction module are provided in SiamOC (Siamese-Occlusion-Correction). They use siamese tracking module to improve the feature extraction ability of the network. The occlusion estimation module  and target location module are based on two assumptions: (i) Different occlusion stat has different depth histogram characteristics; (ii) When the target is occluded temporarily, its speed and movement direction will not be changed significantly. The tracker uses Kalman filter to predict the location of the target considering the speed and direction of target's movement.

However, due to the limited-expression ability of handcrafted features, deep neural networks are introduced to RGBD tracking.
Generally, deep learning-based models follow two principles:
1) Trackers integrate RGB features extracted by pre-trained deep neural networks with handcrafted depth features into heuristic tracking frameworks \cite{2019DAL, zhao2021tsdm}.
2) Trackers are trained on both RGB and depth data jointly to obtain deep depth features as well as deep RGB features.
However, up to now, almost all trackers follow the first principle and remain on using the deep features extracted by pre-trained models on RGB datasets.
Some representative models are briefly introduced here.
%SiamOC \cite{zhang2020occlusion} provides an occlusion estimation module and target location correction module in addition to Siamese tracking network.
%The occlusion estimation module estimate the occlusion state of the target by depth histogram characteristics.
%assumes different objects have different depth histogram characteristics.
%While, the target location module is based on the assumption that, when the target is occluded temporarily, its speed and movement direction will not be changed significantly.
In addition to the Siamese tracking network, SiamOC \cite{zhang2020occlusion} provides two modules to consider both depth histogram characteristics and movement smoothness, simultaneously.
The underlying assumption is that different objects have different depth histogram characteristics.
DAL \cite{2019DAL} embeds depth information into deep features through the reformulation of a deep discriminative correlation filter (DCF).
TSDM \cite{zhao2021tsdm} equips SiamRPN++ \cite{li2019siamrpn++} with two assistant depth-related modules.
% \textbf{DAL \cite{2019DAL}.} In Depth-aware Long-term Tracker, the depth information is embedded into deep features through the reformulation of deep discriminative correlation filter (DCF). Depth modulates the original DCF by re-weighting the kernel of correlaiton filter in terms of the depth similarity of the predicted object position. Moreover, the same depth-aware correlation filter is used for object re-detection. With a speed comparable to real-time trackers, DAL obtains a state-of-the-art performance.
Until 2021, the trainable RGBD tracker DeT \cite{yan2021depthtrack} is first proposed with duplicating a separate feature extraction branch for depth colormaps.
% for depth colormaps transferred from depth maps.

% \subsubsection{Heuristic models}

% \noindent
% \textbf{STC \cite{xiao2017robust}}. A hierarchical, two-layered (global-template layer and local-parts layer) RGBD tracker which adaptively fues a variety of features, derived from both RGB and depth images. RGB and depth images features are separately used in KCF to find the target position in the global layer. We extract color attributes \cite{van2009learning} defined as linguistic color labels with eleven basic color terms. In depth feature, we use depth-HOG as depth feature. If an ambiguous situation is detected by global layer, then local layer will enlarge the output region estimated by global layer and use super-pixel segmentation for matching target parts.

% \noindent
% \textbf{OAPF \cite{MESHGI201681}.} Occlusion Aware Particle Filter (OAPF) proposes a particle filter framework with occlusion awareness using a occlusion flag. Based on the "occlusion flag", the particle goes through a feature-based template matching process (no-occlusion case) or branches to an occlusion case in which all of the occluded particles are treated uniformly. They prepared a rich pool of features coming from the depth channel and color channels (Edges feature, HOG feature, Texture feature, Probability mask etc.). For each feature, the likelihood of all the particles. Then, the likelihood of a non-occluded particle was calculated as the product of all of the features to realize robust feature fusion.

% \noindent
% \textbf{DS-KCF \cite{2016DS}.}

\subsection{VOT participants}
Since 2019, there have been 3 VOT-RGBD challenges \cite{kristan2019seventh, Kristan2020a, kristan2021ninth} held annually, consisting of 13 participating trackers in total.
Most participants in VOT just provide a tracking model and brief introduction without specific publications yet they show outstanding performance in multi-challenges.
Therefore, we here review representative VOT participants as well. % as an integrated whole.

% \noindent
% \textbf{ATCAIS.} The proposed tracker combines both instance segmentation and the depth information for accurate tracking. The tracker ATCAIS is based on the ATOM tracker and the HTC instance segmentation method which is re-trained in a category-agnostic manner. The instance segmentation results are used to detect background distractors and to refine the target bounding boxes to prevent drifting. The depth
% value is used to detect the target occlusion or disappearance and re-find the target. The submitted tracker did not report the confidence.

% \noindent
% \textbf{LTDSEd.} The tracker LTDSEd divides each long-term sequence into several short episodes and tracks the target in each episode using short-term tracking techniques. The visibility of the target is judged by the outputs from short-term components. See also the description of LT-DSE from the same authors.

\noindent
\textbf{2019 Winner: SiamDW-D} is a long-term tracker which addresses the problems of target appearance variations and frequent target loss.
It contains three parts, \textit{i.e.}, the main tracker, a re-detection module, and a multi-template matching module.
The main tracker is based on \cite{zhang2019deeper}, and further equips with an online updating model \cite{2016Learning,2019ATOM}.
The re-detection module is triggered when the main tracker is not confident in its predictions.
%However, in some cases, the confidence of main tracker is not reliable for re-detection module triggering, therefore a
The multi-template matching module is to output a more reliable estimation when the tracking results are unreliable with history templates.
Moreover, depth information is used to estimate the disappearance of target objects.

\noindent
\textbf{2020 Winner: ATCAIS} combines both instance segmentation and depth information for accurate tracking.
It is based on ATOM~\cite{2019ATOM} and the HTC instance segmentation method~\cite{chen2019hybrid}, which is retrained in a category-agnostic manner.
The instance segmentation results are used to detect background distractors and to refine the target bounding boxes to prevent drifting.
The depth value is used to detect the target occlusion or disappearance and redetect the target.

\noindent
\textbf{2021 Winner: STARK\_RGBD} is a method combining STARK~\cite{yan2021learning} and DiMPsuper~\cite{bhat2019learning}. %STARK is a powerful transformer-based tracker with a siamese structure.
Here STRAK variant DeiT~\cite{DeiT} is used to strengthen the features of STARK.
%We notice the STARK method is not good at handling the appearance change of the target.
To better handle the appearance change of the target, DeiT combines with the DiMPsuper model.
Specifically, when the STARK tracker's confidence is low or the prediction of STARK suddenly strays away, DiMPsuper takes over the tracking process, providing an appearance adaptive result.
%When the STARK's confidence resumes, we switch back to STARK. We also design
In addition, a refinement module based on AlphaRefine~\cite{AlphaRefine} is applied to the final output of the whole tracking system for further boosting the quality of box estimation.

Besides the annual winners, other participants also give solutions on RGBD tracking.
In DDiMP\cite{Kristan2020a}, depth information is utilized to prevent scale from changing too quickly.
As targets cannot have very large displacements in two consecutive frames, SiamM\_Ds\cite{kristan2019seventh} averages the depth from the depth image inside the object candidate to determine the depth of the target as a constraint.
TALGD\cite{kristan2021ninth} uses depth images for occlusion or disappearance reasoning and target retrieval.
CLGS-D\cite{kristan2019seventh} uses depth maps to filter region proposals.
Note that some participants do not use depth information indeed, including sttc\_rgbd, DRefine, and STARK\_RGBD, while they still keep high performance.

\begin{figure*}
	\centering
		\includegraphics[width=\linewidth]{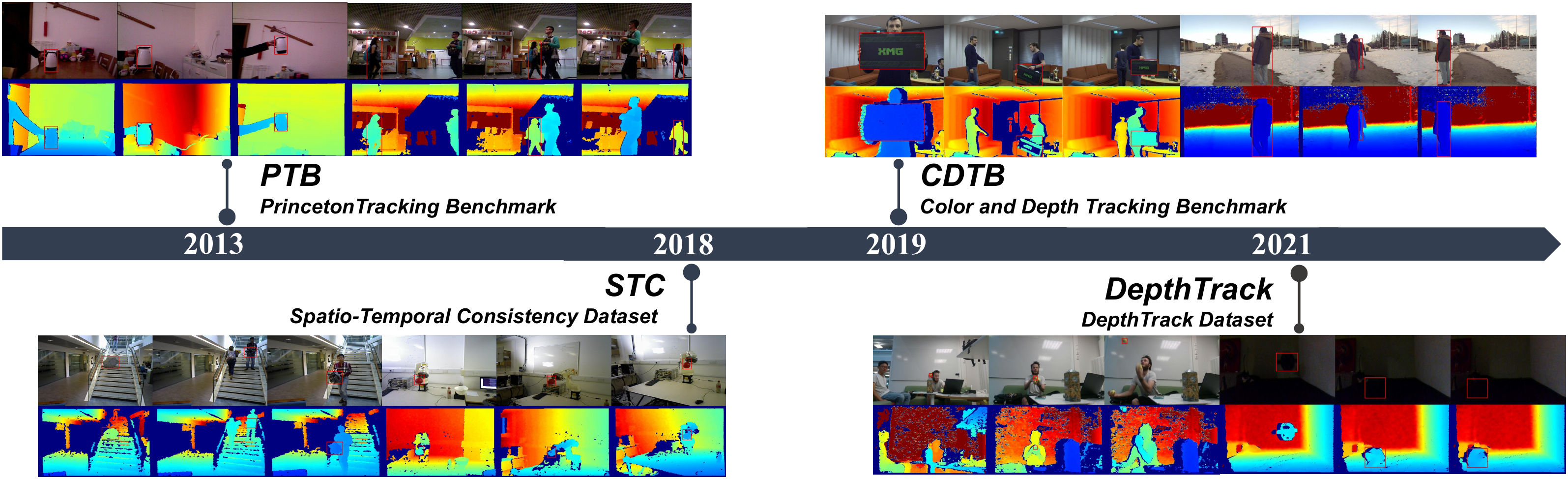}
	\caption{Samples in different RGBD tracking datasets.}
	\label{fig_dataset}
\end{figure*}

\section{Datasets and Evaluation Metrics}\label{dataset_metric}
%statistics

\subsection{Datasets}
In RGBD tracking, early datasets only contain a few sequences for evaluation.
In 2012, a small-scale dataset called BoBoT-D \cite{GM2012Adaptive} was proposed, consisting of five RGBD video sequences captured by Microsoft Kinect v1.0.
In \cite{2014Multi}, four videos were captured and utilized for evaluation with four kinds of objects (Book, Face, Inno, and TeaCan).
These video sequences represent different challenges, such as occlusion, rotation, illumination change, shape variation of a flexible object, and small target, and are manually annotated the groundtruth every five frames.

From 2013, benchmark datasets have appeared for wide use.
Up to now, there are four datasets dedicated designed and widely used for RGBD object tracking.
The Table \ref{tbl_dataset} compares the four representative RGBD tracking datasets.
Fig.~\ref{fig_dataset} gives chronology and examples of these datasets.
The details of each dataset are given as follows:

\noindent
\textbf{Princeton Tracking Benchmark (PTB) \cite{song2013tracking}.} As the first dataset designed for RGBD object tracking, PTB contains 100 RGBD video clips recorded by Microsoft Kinect v1.0.
There are only 3 types: human, animal, and rigid object, with more than half of the sequences, are for people tracking.
For a fair comparison, PTB withholds groundtruth for 95 videos and hosts an online evaluation server to allow new result submissions.
The remaining 5 videos are public for tracker validation before submitting their results.
It is worth noting that the RGB and depth channels are poorly calibrated in PTB.
In approximately 14\% of sequences, the RGB and D channels are not synchronized and approximately 8\% are miss-aligned, which was fixed by \cite{Bibi_2016_CVPR}.

\noindent
\textbf{Spatial-Temporal Consistency Dataset (STC) \cite{xiao2017robust}.} It was proposed in 2018 to address the drawbacks of the PTB dataset.
It is designed to increase the data diversity with a compact size of only 36 sequences for short-term RGBD tracker evaluation.
STC dataset constrains the dataset mostly to indoor scenarios and there are only a few low-light outdoor video clips.
Although the STC dataset is the smallest dataset among the RGBD tracking family, it is annotated with 13 attributes and uses two kinds of evaluation metrics imported from OTB \cite{OTBPAMI2015} and VOT protocols \cite{2016The}.

\noindent
\textbf{Color-and-Depth Tracking Benchmark (CDTB) \cite{Lukezic_2019_ICCV}.} It includes 80 video sequences for long-term tracking with an average video length of 1274 frames.
With long-term settings, objects are possibly fully occluded or out-of-view for a long duration and thus, it can be used to evaluate the re-detection performance.
Note that the CDTB is the only dataset acquired by multiple color-and-depth sensors, which guarantees its diversity of realistic depth signals.
Indoor, as well as outdoor scenarios, are covered to extend the tracking domains.

\noindent
\textbf{DepthTrack Dataset \cite{yan2021depthtrack}.} It consists of 200 sequences, which is the currently largest and most diverse dataset for RGBD tracking.
Specifically, it includes the most diverse object types (46 categories in 50 test sequences), scenarios (indoors and outdoors with 15 attributes), and video length (varying from 143 to 3816 frames).
Until now, it is the first and the only RGBD tracking dataset divided into training and test sets.
In addition, with long-term tracking settings, the DepthTrack dataset is dedicated to exploring the depth-related power to assist in tracking challenges.

\begin{table*}[!t]
\caption{Comparison of existing RGBD tracking datasets. ``LT/ST'' denotes long-term or short-term sequences. }
\centering
\label{tbl_dataset}
\begin{tabular}{|c|c|c|c|c|c|c|c|c|c|c|}
\hline
Dataset  & Publ. & LT/ST & \#Seq. & \#Frame & \#Avg.length& \#Attr. & \#Split & Scenario & Sensor & Resolution\\
\hline
PTB \cite{song2013tracking}  & CVPR & LT & 100 & 20,332 &203&5&- & indoor & Kinect & $640\times480$\\
STC \cite{xiao2017robust}  & TCYB & ST & 36 & 9,195 &255 &12 &-& indoor \& outdoor &Xtion& $640\times480$\\
CDTB \cite{Lukezic_2019_ICCV}  & ICCV & LT & 80 & 101,956 & 1274 &13 &- & indoor \& outdoor &Kinect; Basler& $960\times540$\\
DepthTrack \cite{yan2021depthtrack} & ICCV & LT & 200 &294,600 &1473 &15&150/50& indoor \& outdoor& RealSense &$640\times360$\\
\hline
\end{tabular}
\end{table*}

\subsection{Evaluation metrics}
Although there are only four datasets, their evaluation protocols are different. In this section, we give representative evaluation metrics for RGBD object tracking in detail, \textit{i.e.}, Success Rate (SR),  Pr-Re (Precision-Recall), and F-score.

\noindent
\textbf{Success Rate (SR).}
Inspired by PASCAL VOC chllenge \cite{2009Pascal}, for $t$-th frame, the overlap ratio $r_t$ between the predicted bounding box $A_{t}$ and the groundtruth bounding box $G_{t}$ is:
\begin{equation}
\small
r_t =\left\{
\begin{array}{rcc}
& \frac{area(A_{t} \cap G_{t})}{area(A_{t} \cup G_{t})} & %\qquad
both~A_{t}~and~G_{t}~exist \\
& 1 &neither~A_{t}~or~G_{t}~exists \\
& -1 &otherwise
\end{array}
\right.
\end{equation}
Then, a minimum overlapping area ratio $r$ can be used to decide whether the output is correct.
Thus, the average success rate $R$ of each tracker is defined as follows:
\begin{equation}
R = \frac{1}{N}\sum^{N}_{t=1}{u_t}, \qquad where~u_t = \left\{ \begin{aligned} &1
~if~r_t>r\\&0~otherwise \end{aligned} \right.
\end{equation}
where $u_t$ denotes whether the output bounding box of the $t$-th frame is acceptable, and $N$ is the number of frames.

% \noindent
% \textbf{Overlap Score.} Given the tracked bounding box $b_t$ and the groundtruth bounding box $b_a$, the overlap score is defined as $S = \frac{b_t \cap b_a}{b_t \cup b_a}$, which is indeed the intersection and union of two regions.

%\noindent
%\textbf{Accuracy \& Failure.} Traditional VOT short-term challenges evaluate the

\noindent
\textbf{Precision-Recall (Pr-Re) \& F-score.} According to the settings of the VOT challenges~\cite{kristan2019seventh}\cite{Kristan2020a}, the most popular metric in RGBD tracking is the precision-recall and F-score \cite{9054960}. As the video length over different datasets and sequences varies dramatically, there are frame-based and sequence-based evaluation protocols.
At frame $t$, $\theta_t$ is a prediction confidence score and $\tau_{\theta}$ is a classification threshold.
If the predicted confidence score $\theta_t$ is not below $\tau_{\theta}$, $A_{t}(\tau _{\theta })$ is used to denote the corresponding prediction.
%If the predicted confidence score $\theta_t$ at frame $t$ is below $\tau_{\theta}$.
Otherwise, the output is an empty set and we set $A_{t}(\tau _{\theta })=\varnothing$.
Thus, $\Omega (A_{t}(\tau _{\theta }),G_{t})$ can be used to indicate the intersection-over-union (IoU) between the prediction result $A_{t}(\tau _{\theta })$ and the groundtruth $G_{t}$.
Then, the {\em frame-based evaluation} is as follows:
\begin{equation} \label{eq:pr_re_f_frames_based}
\begin{split}
Pr(\tau _{\theta }) &= \frac{1}{N_{p}} \sum_{A_{t}(\tau _{\theta })\neq \varnothing } \Omega (A_{t}(\tau _{\theta }),G_{t}), \\%\qquad
Re(\tau _{\theta }) &= \frac{1}{N_{g}} \sum_{G_{t}\neq \varnothing }  \Omega (A_{t}(\tau _{\theta }),G_{t}), \\
F(\tau _{\theta })  &=\frac{2Re(\tau _{\theta })Pr(\tau _{\theta })}{Re(\tau _{\theta })+Pr(\tau _{\theta })},
%{Fscore} &= Max(F(\tau _{\theta }) ),
\end{split}
\end{equation}
%$G_{t}$ denotes the groundtruth of the target and
where $F(\tau _{\theta })$, $Pr(\tau_{\theta})$ and $Re(\tau_{\theta})$ denote the F-score metric, the precision (\textit{Pr}) and the recall (\textit{Re}) over all frames, respectively.
$N_{p}$ denotes the number of frames in which the target is predicted visible, and $N_{g}$ denotes the number of frames in which the target is indeed visible.

For {\em sequence-based evaluation}, the precision-recall over all sequences is as follows:
\begin{equation} \label{eq:pr_re_f_sequences_based}
\begin{split}
 Pr(\tau _{\theta }) &= \frac{1}{M} \sum_{i=1}^{M} Pr^{i}(\tau_{\theta}),\\%\qquad
Re(\tau _{\theta }) &= \frac{1}{M} \sum_{i=1}^{M} Re^{i}(\tau_{\theta}), \\
%F(\tau _{\theta })  &=\frac{2Re(\tau _{\theta })Pr(\tau _{\theta })}{Re(\tau _{\theta })+Pr(\tau _{\theta })},
\end{split}
\end{equation}
where $Pr^{i}(\tau_{\theta})$ and $Re^{i}(\tau_{\theta})$ denote the precision and recall metrics for $i$-th sequence among $M$ test videos.
F-score is obtained in the same way as Eq.~\ref{eq:pr_re_f_frames_based}.
%The confidence threshold is denoted as $\tau_{\theta}$.

\begin{figure}[tbp]
\centering
\subfigure[STC]{
\begin{minipage}[t]{0.29\linewidth}
\centering
\includegraphics[width=1in]{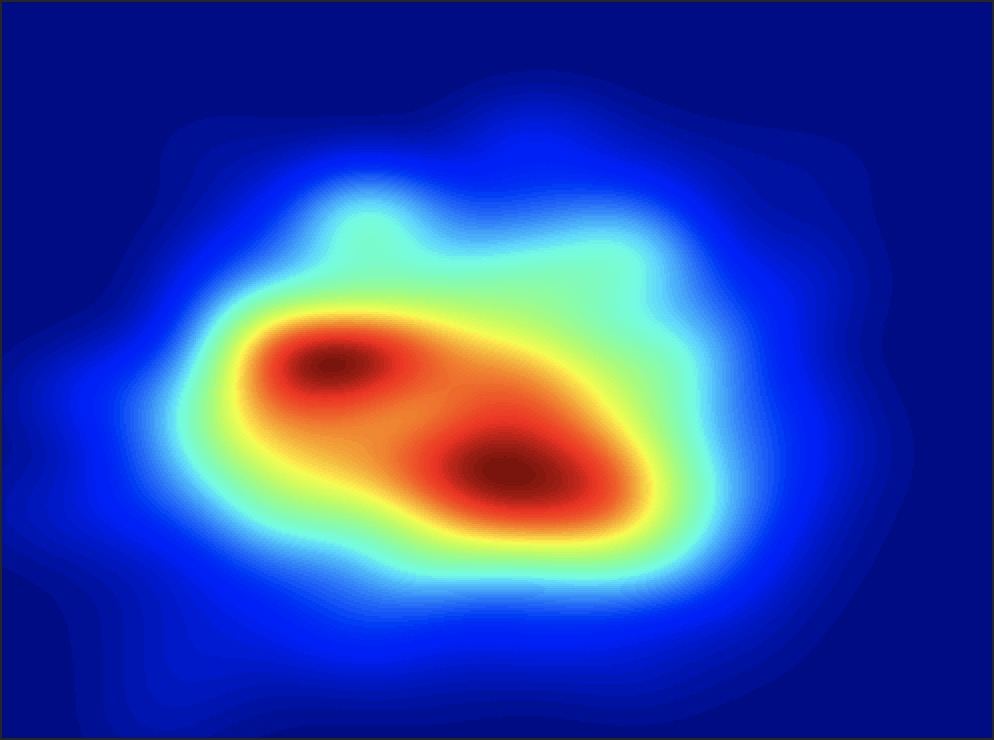}
\centering
\end{minipage}
}
\subfigure[CDTB]{
\begin{minipage}[t]{0.29\linewidth}
\centering
\includegraphics[width=1in]{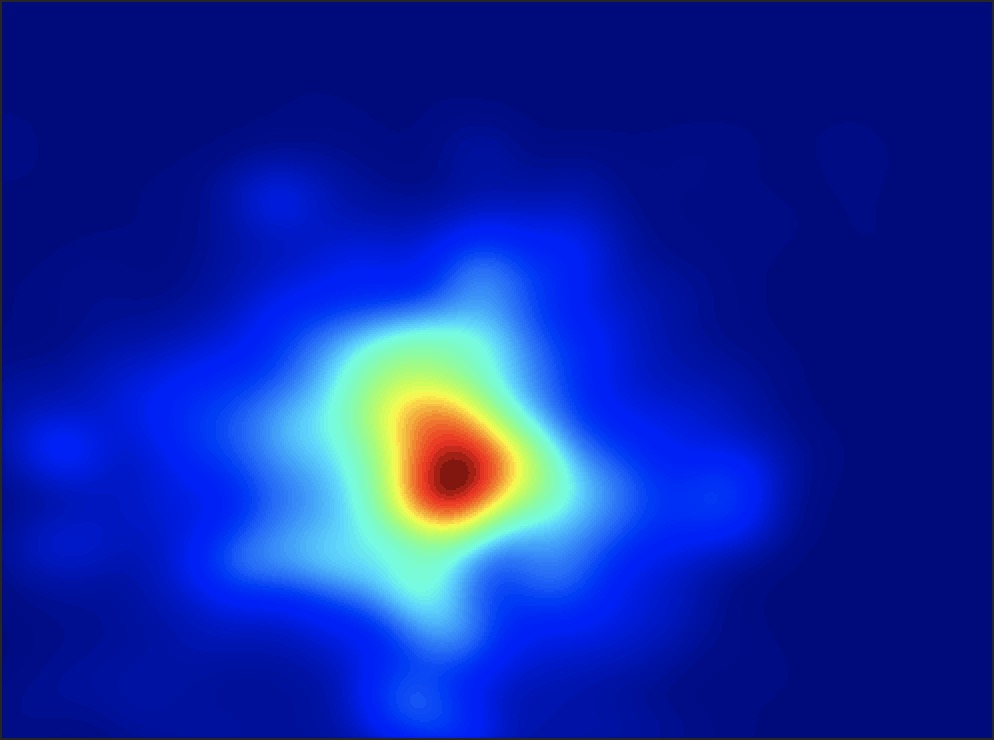}
\centering
\end{minipage}
}
\subfigure[DepthTrack]{
\begin{minipage}[t]{0.29\linewidth}
\centering
\includegraphics[width=1in]{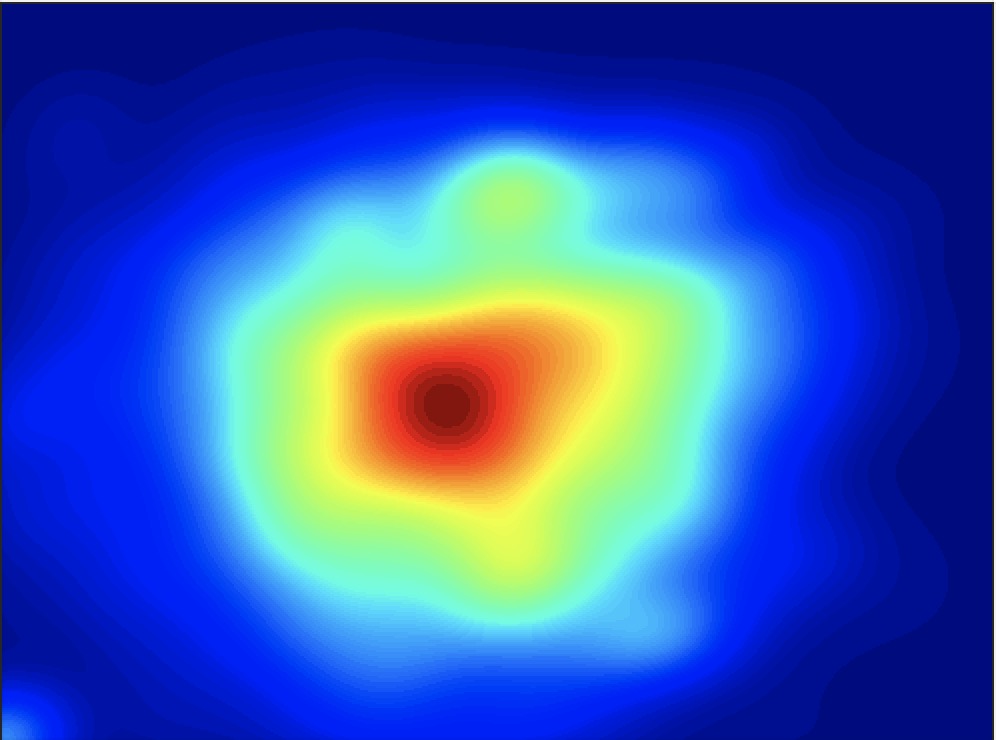}
\centering
\end{minipage}
}
% \centering
\caption{Target center distributions in RGBD datasets.}
\label{annotation_distribution}
\end{figure}

\subsection{Annotation and Attribute}
All datasets are given with per-frame axis-aligned bounding box annotation.
% The unified annotation principle is to select the minimal bounding box which fully contains the target object.
Fig.~\ref{annotation_distribution} illustrates the bounding box distributions for STC, CDTB, and DepthTrack datasets.
Due to its privacy of test data groundtruth, the PTB dataset is not included.
As shown, there are high discrepancies between bounding box distributions of different datasets.

In the field of object tracking, ``attribute'' defines possible challenging factors and different characteristics of each sequence.
Existing datasets propose various attributes from various perspectives.
1) PTB dataset evaluates trackers on \textit{target type (human/animal/rigid), target size (large/small), movement (slow/fast), occlusion (yes/no), and motion type (active/passive)}.
2) STC dataset annotates the attributes per sequence, including \textit{Illumination Variation (IV), Color/Depth Distribution Variation (CDV/DDV), Surrounding Depth/Color Clutter (SDC/SCC), Background Color/Shape Camouflages (BCC/BSC), Partial Occlusion (PO), and Depth/Scale Variation (DV/SV)}.
3) CDTB and DepthTrack datasets share the most attributes, including the common challenges that appeared in RGB-based tracking: \textit{Aspect Change (AC), Fast Motion (FM), Full Occlusion (FO), Non-rigid Deformation (ND), Out-of-plane Rotation (OP), Out-of-frame (OF), Partial Occlusion (PO), Size Change (SC), and Similar Objects (SO)},
and depth-related challenges: \textit{Dark Scene (DS), Depth Change (DC), and Reflective Targets (RT)}.
4) In particular, the DepthTrack dataset considers two more challenges, such as \textit{Background Clutter (BC)} and \textit{Camera Motion (CM)} which are indirectly related to depth scenarios.

\section{Benchmarking and Empirical Analysis}\label{experiments}
To give empirical analysis on the key challenges in RGBD tracking, we specifically conduct a series of studies to better understand the benefits and deficiencies of current arts.
To this end, we first collect the public RGBD tracking data to make a unified benchmark, and then, the experiments conducted on it can uncover meaningful findings on current arts.
Specifically, we propose a depth data quality study to investigate its effect on tracking performance (Sec.~\ref{quality}).
Secondly, we explore depth favorable scenarios which significantly affect the role of depth cues in the tracking process (Sec.~\ref{favorable}).
Then, we analyze the effect of long-term components and compare the speed over various RGBD trackers.
Finally, we quantitatively evaluate the tracking robustness of RGBD trackers against input perturbations (Sec.~\ref{robustness}).
%For unified comparison, we select $F-score$ as the evaluation protocol.

%\subsection{\lizhe{Experiments Setup}}

%\subsubsection{Qualitative results}
%\subsubsection{Discussion}

%\subsection{In-depth Empirical Analysis}\label{indepth}
\begin{table}[!t]
\caption{Overall performance on the hybrid dataset. The top 3 results are shown in {\color{red}red}, {\color{green}green}, and {\color{blue}blue}.
Speed in FPS (frame per second).
}
\centering
\label{tbl_overall}
\begin{tabular}{|l|c|c|c|c|c|}
\hline
% Method &Pr & Re& Fscore  \\
 Method  & Year &\multicolumn{1}{c|}{Pr} & \multicolumn{1}{c|}{Re} & \multicolumn{1}{c|}{F-score} & Speed\\
\hline
DSKCF \cite{camplani2015real}& 2015 & 0.038 & 0.039 & 0.0391 &4.2\\
DSKCF\_shape \cite{2016DS} &2016 & 0.039 & 0.040 & 0.0397 &9.5\\
CA3DMS \cite{liu2018context} &2018 & 0.241 &0.247 &0.244 &17.9\\
CSR\_RGBD++ \cite{Kart_ECCVW_2018} & 2018 &0.159 &0.167 &0.163 & 0.4\\
% OTR \cite{Kart_CVPR_2019} & 0.352 & 0.330& 0.327\\
LTDSEd \cite{kristan2019seventh} &2019& 0.501 & 0.435 & 0.466 &5.7\\
SiamDW-D \cite{kristan2019seventh} &2019& 0.467 & 0.354 & 0.457 &3.8\\
Siam\_LTD \cite{Kristan2020a} &2020& 0.530 & 0.409 & 0.466 &13.0 \\
ATCAIS \cite{kristan2019seventh} &2020&0.571 &\color{red}{0.598} &0.584&1.3 \\
DAL\cite{2019DAL} &2020 &0.653 & 0.496 & 0.526 & \color{green}{21.26} \\
CLGS-D \cite{Kristan2020a} &2020& \color{green}{0.684} & 0.447 & 0.567&7.3 \\
DDiMP \cite{Kristan2020a} &2020& 0.616 & \color{blue}{0.521} & \color{blue}{0.592} &4.8 \\
DRefine \cite{kristan2021ninth} &2021 & 0.614 & 0.506 & 0.590 &4.9\\
SLMD \cite{kristan2021ninth} & 2021& 0.611 & 0.515 & 0.586 & 2.4\\
sttc\_rgbd \cite{kristan2021ninth} &2021& 0.655 & 0.592 & 0.591 & 16.2\\
TSDM \cite{zhao2021tsdm} & 2021 & 0.488 & 0.457 & 0.472 & \color{blue}{19.0}\\
DeT \cite{yan2021depthtrack} & 2021  & 0.586 & \color{green}{0.545} & 0.560 & \color{red}{26.78} \\
TALGD \cite{kristan2021ninth} &2021& \color{blue}{0.663} & 0.517 & \color{green}{0.622} & 1.3\\
STARK\_RGBD \cite{kristan2021ninth} & 2021& \color{red}{0.707} & 0.454 & \color{red}{0.657} & 5.4 \\
% SiamM\_Ds \cite{kristan2019seventh} & \jinyu{0.068} & \jinyu{0.062} & \jinyu{0.065} \\
\hline
\end{tabular}
\end{table}

\begin{figure*}
	\centering
		\includegraphics[width=0.9\linewidth]{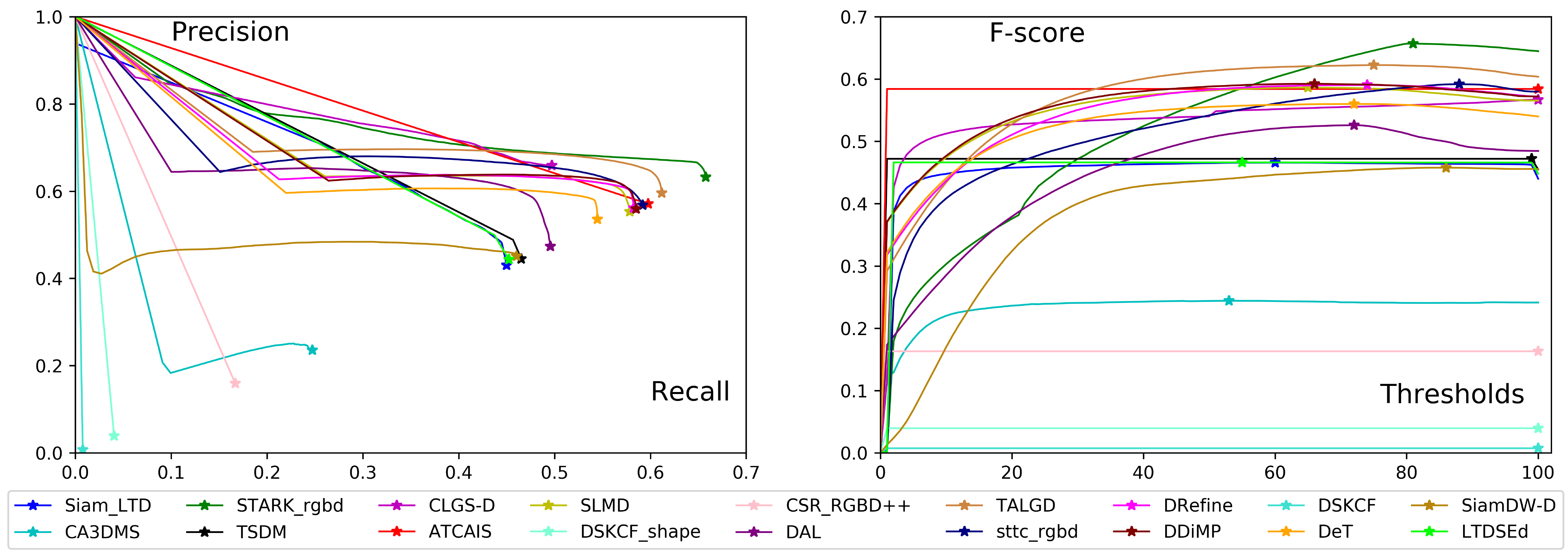}
	\caption{Overall performance of RGBD trackers on the hybrid dataset.}
	\label{fig_overall}
\end{figure*}

\begin{figure}
	\centering
		\includegraphics[width=0.95\linewidth]{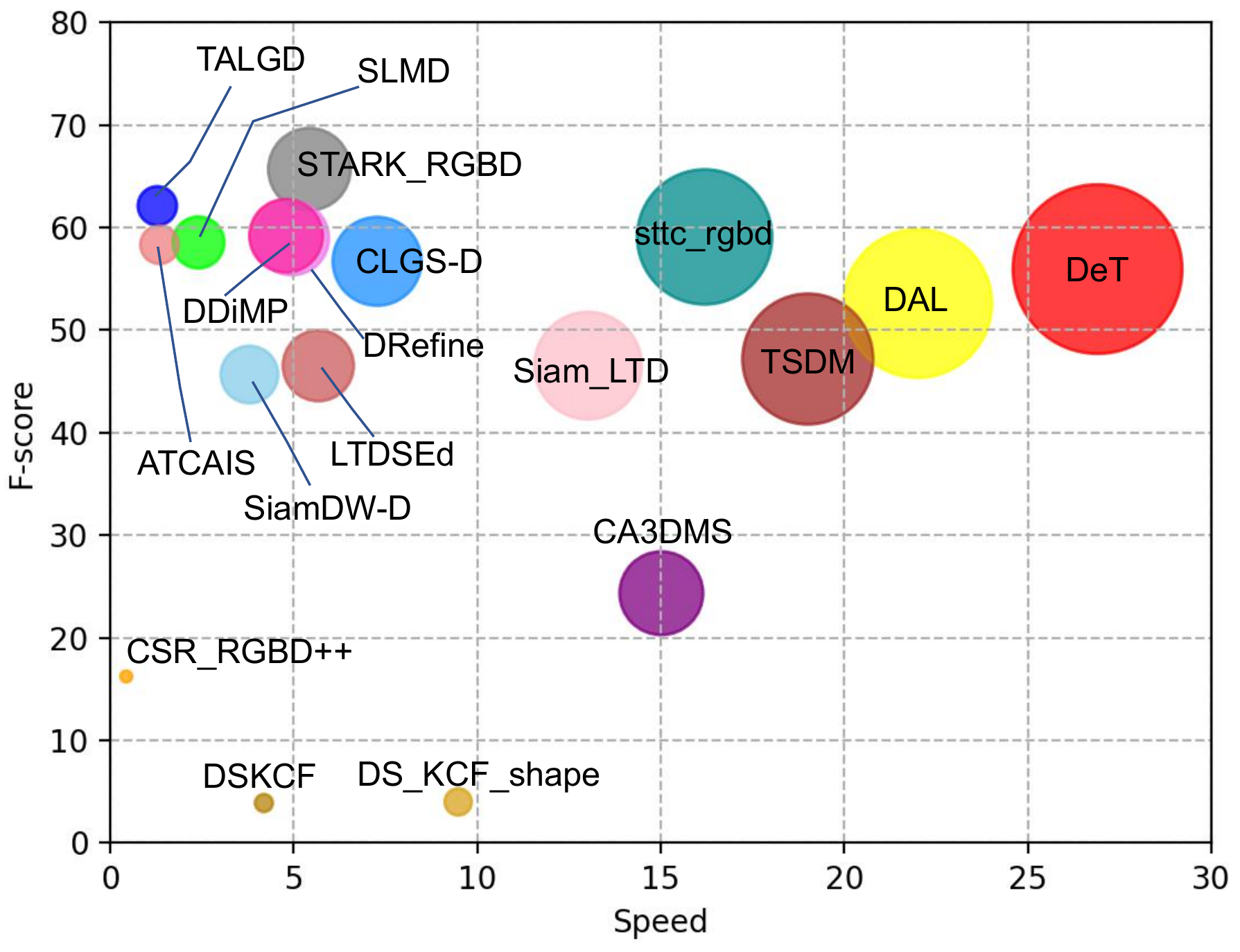}
	\caption{Speed-performance trade-off comparison. We visualize the F-scores with respect to the Frames-Per-Second (fps) tracking speed.
	The circle size is proportional to both the F-score and tracking speed. The larger, the better.}
	\label{fig_bubble}
\end{figure}

\subsection{Benchmarking performance}\label{compare}
%\subsubsection{Quantitative results}
% Table \ref{tbl:experiments_on_DeTrack} shows the performances of \textcolor{red}{24} RGBD trackers on four existing datasets.
For benchmarking performance on different datasets, we
% generate
conduct a hybrid dataset for in-depth analysis.
We collect the public RGBD tracking test data for evaluation from the aforementioned STC \cite{xiao2017robust}, CDTB \cite{Lukezic_2019_ICCV}, and DepthTrack \cite{yan2021depthtrack} datasets.
Thus, the hybrid dataset contains 166 video sequences, with 187,514 frames. %\jinyu{?} sequences' length over 1,000 frames.
For a unified comparison, we use a frame-based F-score (see Eq.~\ref{eq:pr_re_f_frames_based}) as the evaluation protocol.
All the benchmark trackers are representative and have officially available implementations.
All results are obtained by running with their official codes.
It is worth noting that some early trackers are not able to report confidence scores so we set the confidence score as 1 for every frame.

Overall performance is given in Table~\ref{tbl_overall} and Fig.~\ref{fig_overall}.
Obviously, data-driven models greatly outperform conventional heuristic ones.
Traditional methods perform poorly on current larger and more challenging datasets.
From 2015 to 2018, the development of RGBD trackers was slow.
In fact, the VOT RGBD competition which was first held in 2019, started to promote the research of RGBD trackers into the field of deep models.
Specifically, VOT participants mostly outperform on F-score, showing high generalization ability and tracking performance.
But their speeds are relatively low, far below the real-time requirement.
%Advanced deep RGBD trackers win on tracking speed,
Here we also report the speed-performance trade-off in Fig.~\ref{fig_bubble}, which has not been included and analyzed in most related works.
As shown, DeT~\cite{yan2021depthtrack} achieves the best on speed-performance trade-off.
However, many high-performance RGBD trackers are sub-optimal, indicating they may sacrifice tracking speed for the sake of tracking accuracy.

% Since visual object tracking is indeed a real-time application, tracking speed is of vital in RGB tracking in a long time.
% However, it is not included and analysed by most RGBD tracking benchmarks.
% Here we report the speed of trackers.

% \begin{table*}%[!t]
% \caption{Data quality. }
% \centering
% \label{tbl_quality}
% \begin{tabular}{|c|c|c|c|c|c|c|c|c|c|c|c|c|c|c|c|c|c|c|c|c}
% \hline
% Depth quality &DeT \cite{yan2021depthtrack} &OTR \cite{Kart_CVPR_2019} & DAL\cite{2019DAL } &TSDM & DDiMP &SiamDW\_D  & LTDSEd & Siam-MDs & DRefine & Ca3dms & CLGSD & STARK\_rgbd & Siam\_LTD & ATCAIS & SLMD & CSRDCF-D & TALGD & sttc_rgbd\\
% \hline
% Low    & 0.621 & 0.407 & 0.452 & 0.421 & 0.512 & 0.391 & 0.407 & 0.048 & 0.505 & 0.356 & 0.446 & 0.588 & 0.402 & 0.568 & 0.490 & 0.141 & 0.581 & 0.544\\
% Medium & 0.666 & 0.337 & 0.460 & 0.398 & 0.538 & 0.401 & 0.406 & 0.040 & 0.520 & 0.310 & 0.486 & 0.639 & 0.418 & 0.573 & 0.520 & 0.138 & 0.625 & 0.524\\
% High   & 0.637 & 0.318 & 0.499 & 0.422 & 0.557 & 0.468 & 0.482 & 0.059 & 0.541 & 0.258 & 0.512 & 0.608 & 0.420 & 0.529 & 0.560 & 0.133 & 0.549 & 0.544\\
% \hline
% \end{tabular}
% \end{table*}

\begin{figure*}
	\centering
		\includegraphics[width=\linewidth]{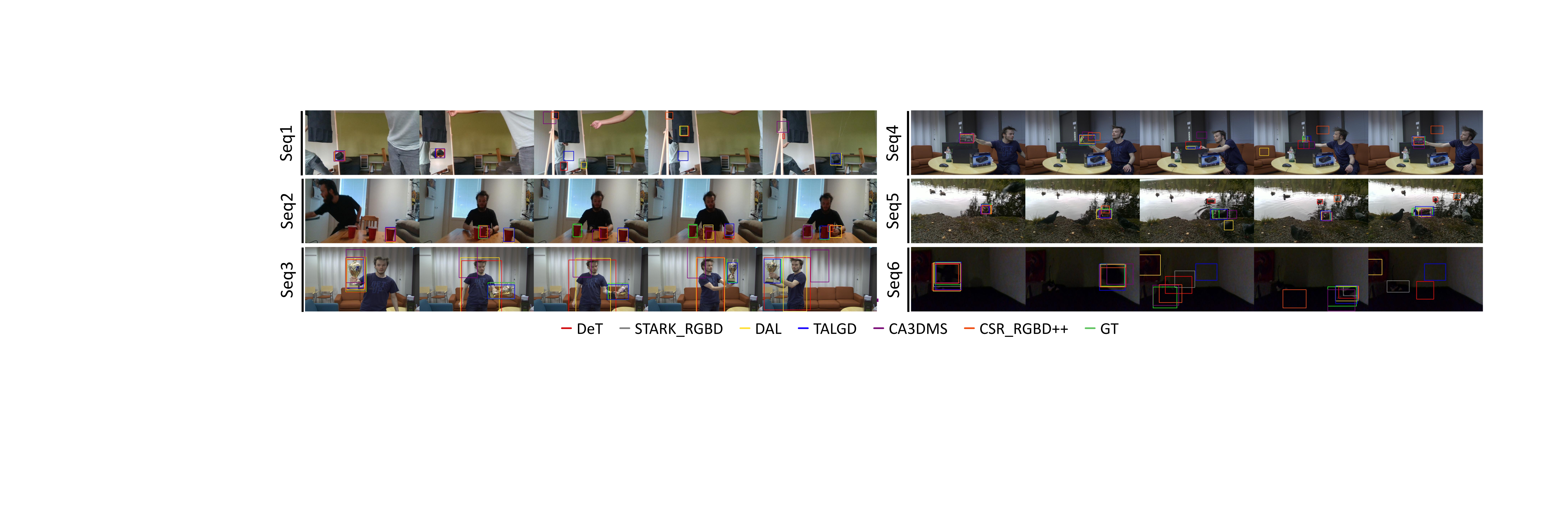}
	\caption{Visualized examples with different challenges. Seq 1: similar objects and background clutter; Seq 2: similar objects; Seq 3: reflective targets; Seq 4: depth change and occlusion; Seq 5: outdoor scenario and similar objects; Seq 6: dark scene.}
	\label{fig_LT}
\end{figure*}

\begin{table}%[!t]
\caption{Depth data quality evaluation. ``Change'' shows the F-score change (in percent) compared to high depth quality ones. The worst 3 results of Change are shown in {\color{red}red}, {\color{green}green}, and {\color{blue}blue}.}
\centering
\label{tbl_quality}
\begin{tabular}{|l|c|c|c|}
\hline
Method &Low/Change& Medium/Change& High\\
\hline
DAL\cite{2019DAL} & 0.351/-21.0\%& 0.419/-5.63\% & 	0.444\\
CA3DMS \cite{liu2018context} & 0.166/-20.2\% &0.205/-1.44\% & 0.208 \\
CSR\_RGBD++\cite{Kart_ECCVW_2018} &0.091/\color{green}{-27.8}\% & 0.099/\color{red}{-21.4}\%& 0.126 \\
DSKCF \cite{camplani2015real}&	0.028/-4.81\% &0.027/\color{green}{-7.22}\% &  0.0291 \\
DSKCF\_shape \cite{2016DS}&0.022/-12.6\% &0.024/-0.81\% &0.0247 \\
% OTR \cite{Kart_CVPR_2019} & 0.407 & 0.337& 0.318 \\
DRefine\cite{kristan2021ninth}  & 0.394/-15.3\% & 0.452/-2.80\% & 0.465 \\
SLMD \cite{kristan2021ninth} &0.373/-22.1\% & 0.459/-4.18\% & 0.479 \\
TALGD\cite{kristan2021ninth}  & 0.516/-0.19\% & 0.510/-1.35\% & 0.517\\
CLGSD \cite{Kristan2020a}& 0.309/\color{red}{-27.8}\% & 0.398/-7.01\% & 0.428  \\
DDiMP \cite{Kristan2020a} & 0.416/-14.7\%  & 0.474/-2.87\%& 0.488  \\
Siam\_LTD \cite{Kristan2020a}& 0.324/-22.8\%  & 0.418/-0.48\% & 0.420  \\
LTDSEd \cite{kristan2019seventh}& 0.370/-17.9\%& 0.434/-3.77\% &  0.451\\
Siam\_LTD  \cite{kristan2019seventh}&0.276/\color{blue}{-25.4}\% & 0.353/-4.59\% &0.370 \\
SiamDW\_D \cite{kristan2019seventh}&  0.377/-16.2\% & 0.427/-5.11\%  & 0.450 \\
TSDM \cite{zhao2021tsdm}  & 0.296/-20.0\%& 0.346/\color{blue}{-6.49}\%& 0.370 \\
ATCAIS \cite{Kristan2020a}& 0.473/-4.25\% & 0.485/-1.82\% & 0.494 \\
STARK\_RGBD\cite{kristan2021ninth} & 0.534/-4.13\%& 0.552/-0.90\%& 0.557 \\
sttc\_rgbd\cite{kristan2021ninth} & 0.427/-10.5\%& 0.455/-4.61\% &  0.477\\
DeT \cite{yan2021depthtrack}  &  0.490/-2.39\% &  0.506/+0.80\% & 0.502 \\

% Siam-MDs  & 0.048& 0.040 & 0.059\\

\hline
Average & 0.328/-15.3\% & 0.371/-4.30\%	& 0.384 \\
\hline
\end{tabular}
\end{table}

% depthquality
\begin{figure}
	\centering
		\includegraphics[width=\linewidth]{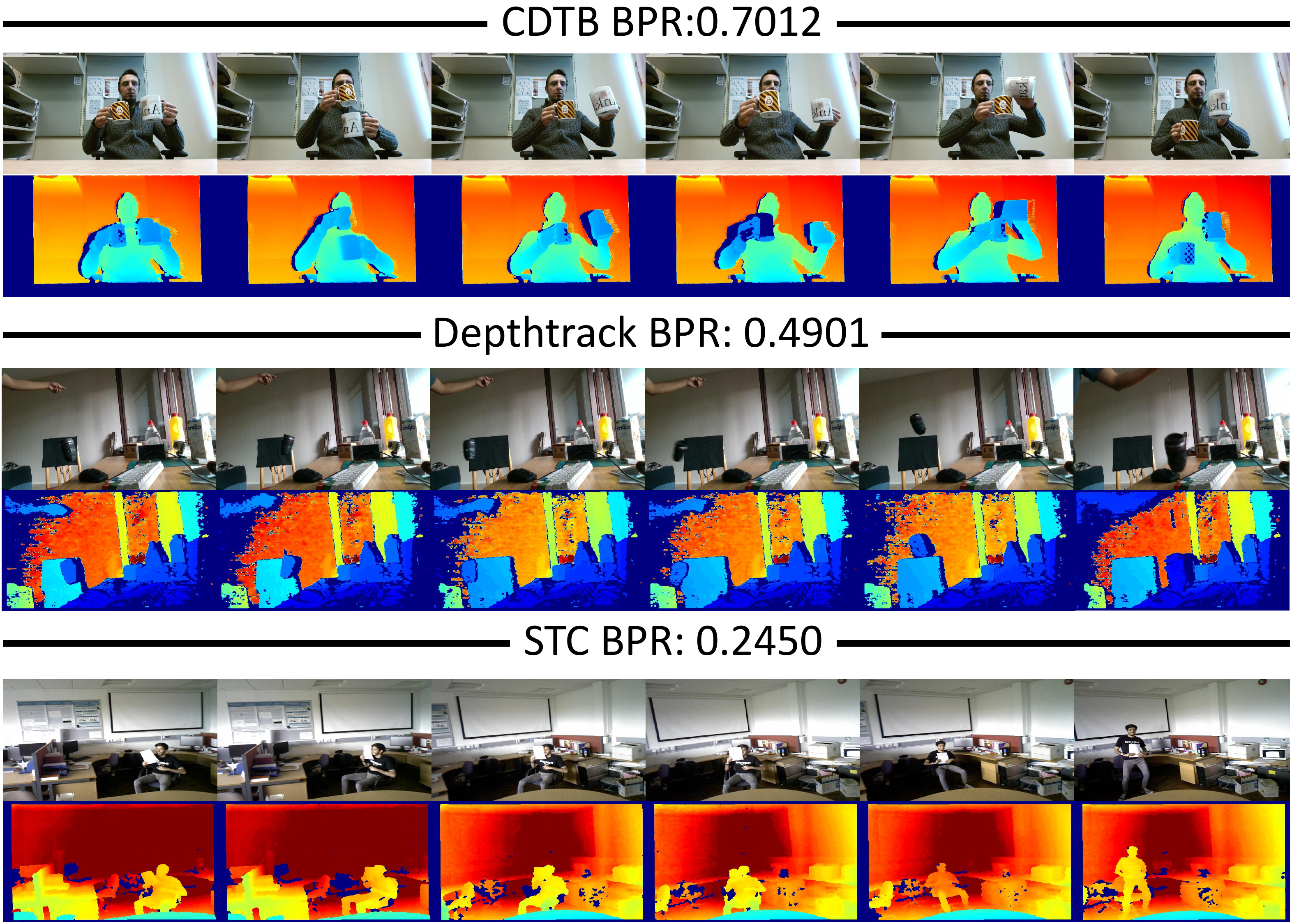}
	\caption{Visualized example sequences with different-level depth qualities.
	The lower BPR, the better depth quality.}
	\label{fig_depthQ}
\end{figure}

\subsection{Effect of depth data quality}\label{quality}
%As we mentioned in Sec.~\ref{dataset_metric}, high-quality and large-scale depth data remains a limitation to the development of RGBD tracking, where depth data quality can be influenced by the depth sensor selection and capturing environments.
Since depth images provide very important complementary information for object tracking, their quality is very important.
Due to the limitations of depth sensors,
the poor depth quality contains challenges including asynchrony between color and depth channels, low resolution, median and Gaussian blur, and color distortion.
To investigate how depth data quality affects tracking performance, we divide the RGBD video sequences into three groups (low quality, medium quality, and high quality) according to their depth quality level for performance comparison.
Hence, we evaluate the Depth Quality (DQ) based on the Bad Point Rate (BPR) proposed in \cite{xiang2015no}, in which $DQ = 1 - BPR$.
BPR is a state-of-the-art no-reference depth assessment metric and matches the texture edges and depth edges to measure the proportion of mismatched pixels.
The lower BPR value denotes the better depth quality.

When evaluating the depth quality of STC, CDTB, and Depthtrack datasets, we calculate the depth quality of the three datasets to be 0.2450, 0.7012, and 0.4901, respectively.
Among them, the depth quality of the CDTB dataset is the worst.
By visualizing the depth map, as shown in Fig.~\ref{fig_depthQ}, %we can find the quality problem.
we find that the resolution of depth images does not match the ones of color images in the CDTB dataset, resulting in lots of empty depth values after aligning the depth image to the color image.
To effectively investigate the quality impact of depth maps, we conduct the experiments by removing these sequences from the CDTB dataset.
%To guarantee the validity of the correlation between depth quality and performance experimentally obtained, we remove the sequences in the CDTB dataset to analyze the performance of the tracking methods.
The results of the tracker performance evaluation on sequences of three depth qualities are shown in Table \ref{tbl_quality}.
The average performance of tracking methods on low-quality depth images drops by 15.27\% compared to high-quality depth images and by 4.30 \% on medium quality depth images.
For traditional methods, the quality of the depth map has a greater impact.
When tested on the low-quality depth maps, these methods have sharp drops in the results.
While, for deep learning methods, the quality of the depth maps has relatively little impact on their performance.
The inherent reason is that most deep models heavily depend on the pre-trained RGB tracking models. % depth map features.
With only heuristic depth features used on specific occasions, \textit{e.g.}, occlusion detection, the depth quality does not affect the final performance so much.
%Thus the depth quality only shows weak impacts on the deep models.
%To some extent, we can see these pre-trained models can reduce the requirement of data quality, especially the size and quality of the data.
Among all these methods, DeT\cite{yan2021depthtrack} conversely shows 0.80\% higher performance on medium-quality datasets relative to high-quality datasets.
%DeT is the baseline proposed in the Depthtrack \cite{yan2021depthtrack}, which is the first work to collect and provide RGBD tracking training datasets.
It can be seen that the quality of DepthTrack dataset is generally low.
It is due to that DeT is trained on the DepthTrack training set, indicating there may be domain overfitting on DepthTrack.

%DeT is also the first method to do pre-training separately on the depth branch, which shows that the deep model is more robust to depth quality problems.

% \jinyu{low-quality camera sensors (including low resolution, low bit depth, low frame rate and color distortion)
% asynchrony between color and depth channels; hole filling; Median and Gaussian Blur. shadow; syn; }
\begin{table}[!t]
\caption{Attribute-based performance evaluation. BC=background clutter, RT=reflective targets, DS=dark Scenes, SO=similar objects, DC=depth change.}
\centering
\label{tbl_attribute}
\renewcommand\tabcolsep{5pt}
\begin{tabular}{|l|l|l|l|l|l|}
\hline
 Method  &\multicolumn{1}{c|}{BC} & \multicolumn{1}{c|}{RT}& \multicolumn{1}{c|}{DS} & \multicolumn{1}{c|}{SO} & \multicolumn{1}{c|}{DC}  \\
%  Method  & BC & RT & DS &SO & DC\\
\hline
DeT \cite{yan2021depthtrack} & \color{green}{0.437} & 0.619 & 0.627 & 0.458 & 0.578\\
TSDM \cite{zhao2021tsdm} &0.251 & 0.562 & 0.564 &0.341 &0.481\\
DAL\cite{2019DAL} &0.334 &0.543 &0.620 &0.446 &0.515\\
CA3DMS \cite{liu2018context}& 0.236 &0.131 &0.316 & 0.199 &0.201\\
CSR\_RGBD++ \cite{Kart_ECCVW_2018}& 0.085 &0.160 &0.186&0.110 &0.145\\
DSKCF\_shape\cite{2016DS} & 0.023 & 0.031 & 0.049 & 0.028 & 0.027\\
STARK\_RGBD \cite{kristan2021ninth} & \color{red}{0.519} & \color{red}{0.776}& \color{red}{0.789} & \color{red}{0.493} & \color{red}{0.699} \\
DRefine \cite{kristan2021ninth} & 0.393 &0.701 &0.720 &0.440 &0.613\\
SLMD \cite{kristan2021ninth} &0.394 &0.651 & 0.719 & \color{blue}{0.464} &0.609\\
TALGD \cite{kristan2021ninth} &\color{blue}{0.431} & \color{blue}{0.721} & \color{green}{0.764} &0.448 & \color{green}{0.683}\\
sttc\_rgbd \cite{kristan2021ninth} & 0.386 & \color{green}{0.724} & 0.711 &0.408 &0.633\\
CLGS-D \cite{Kristan2020a} & 0.283 &0.647 & 0.676 &0.386 &0.630\\
DDiMP \cite{Kristan2020a} &0.379 &0.648 &0.724 & \color{green}{0.466} &0.618\\
Siam\_LTD \cite{Kristan2020a} &0.310& 0.560 & 0.554 &0.327  &0.473\\
ATCAIS \cite{kristan2019seventh} &0.428 & 0.661 & \color{blue}{0.753} &0.436 & \color{blue}{0.652} \\
LTDSEd \cite{kristan2019seventh} &0.366&0.532 &0.633&0.421&0.305\\
% OTR \cite{Kart_CVPR_2019} &\jinyu{0} &0.294 &0.351 &0.307 &0.336 \\
% \jinyu{SiamM\_Ds} \cite{kristan2019seventh}&0.036 &0.082 &0.072&0.050&0.054 \\
SiamDW-D \cite{kristan2019seventh} & 0.345&0.428 &0.637 &0.371 &0.295 \\
\hline
\end{tabular}
\end{table}

\begin{figure}
	\centering
		\includegraphics[width=\linewidth]{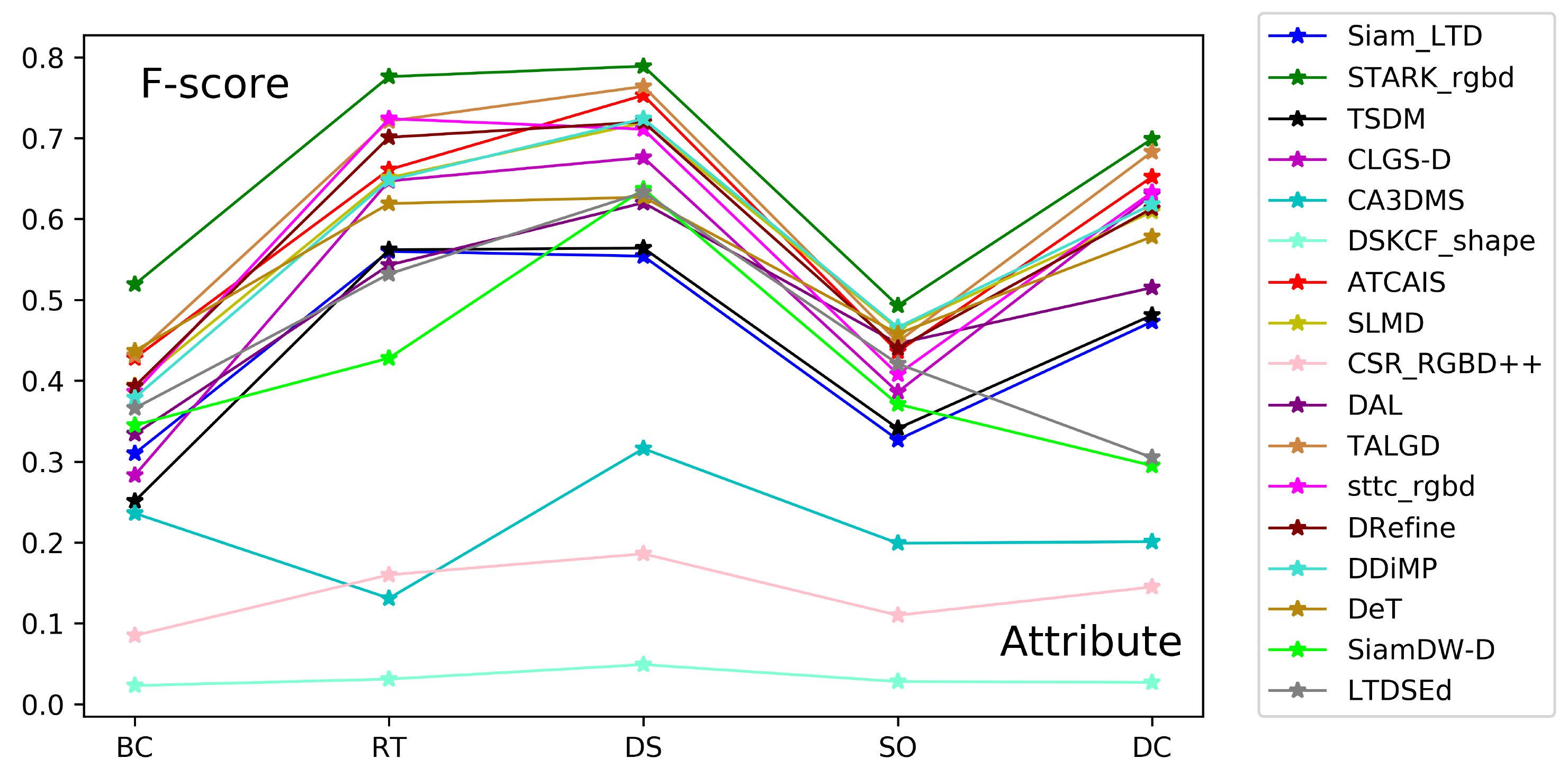}
	\caption{Attribute-based F-scores of compared trackers.}
	\label{fig_attribute}
\end{figure}

\subsection{Attribute-based analysis}\label{favorable}
In RGBD tracking, numerous factors affect the tracker performance.
To investigate how trackers benefit from specific components, we evaluate trackers' performance under different scenarios.
This helps to reveal the strength of RGBD trackers and find out the scenarios that which depth cues can benefit.
Moreover, depth favorable scenarios are essential for researchers to bridge the gap between 2D and 3D tracking, which are surprisingly unexplored in RGBD tracking.
We select and classify the depth-related issues from existing RGBD tracking datasets.
The depth-related scenarios tags include Depth Change (DC), Reflective Target (RT), Dark Scenes (DS), Background Clutter (BC), Similar Objects (SO).
%Illumination Change (IC), Occlusion (OC), and Color Shift (CS).
We specifically evaluate the tracker performance within these attributes.
%, with the related sequences distribution given in Fig. \jinyu{?}.
STARK\_RGBD\cite{kristan2021ninth} performs best on correlated frames with all listed types of attributes, which demonstrates its effectiveness.
As we reviewed in Sec.~\ref{model}, STARK\_RGBD does not use depth information indeed, while it leads the leaderboard of many RGB tracking benchmarks.
Thus, the tracking performance highly relies on the pre-trained trackers' discriminative ability.
Despite this, dark scenes and depth change are well handled by TALGD\cite{kristan2021ninth} by using depth maps for reasoning target disappearance and retrieval.
%{\color{red}why it is better than others for such two challenges?}
DeT\cite{yan2021depthtrack} shows good performance on background clutter, due to its depth deep features which can distinguish the objects from background distractors in depth colormaps.
In addition, DDiMP\cite{Kristan2020a} and sttc\_rgbd\cite{kristan2021ninth} are good at reflective targets and similar objects, respectively.
Overall, background clutter and similar objects are especially challenging for RGBD trackers with the highest F-scores of 0.519 and 0.493, respectively.

\begin{table}[!t]
\caption{Long-term component analysis. ``Drop'' denotes the performance drop after target loss. Short-term trackers are shown in {\color{red}red}, and heuristic methods are shown in {\color{brown}brown}. } %66 sequences 98558 frames }
\centering
\label{tbl_longterm}
\begin{tabular}{|l|c|c|c|c|}
\hline
% Method & LT/ST &Before Occ & After Occ &F-score Drop\\
 Method  &\multicolumn{1}{c|}{LT/ST} & \multicolumn{1}{c|}{Before Occ}& \multicolumn{1}{c|}{After Occ} & \multicolumn{1}{c|}{Drop $\downarrow$}   \\
\hline
DeT \cite{yan2021depthtrack} &\color{red}{ST} &0.731 &0.341 &\color{red}{0.390}\\
CA3DMS \cite{liu2018context} &\color{brown}{LT}  &0.444&0.090 &\color{brown}{0.354}\\
CSR\_RGBD++ \cite{Kart_ECCVW_2018} &\color{brown}{LT}  &0.320 &0.028 &\color{brown}{0.292}\\
Siam\_LTD \cite{Kristan2020a}  &LT & 0.544 &0.315 &0.229\\
ATCAIS \cite{kristan2019seventh} &LT & 0.653&0.457 &0.196\\
DRefine \cite{kristan2021ninth} &\color{red}{ST} &0.646 &0.456 &\color{red}{0.190}\\
TSDM \cite{zhao2021tsdm}&LT  &0.530 &0.351 &0.179\\
SiamDW-D \cite{kristan2019seventh}&LT & 0.496 &0.318 &0.178\\
sttc\_rgbd \cite{kristan2021ninth} &\color{red}{ST} &0.655 &0.483 &\color{red}{0.172}\\
CLGS-D \cite{Kristan2020a} &LT  &0.641&0.470 &0.171\\
DDiMP \cite{Kristan2020a}  &\color{red}{ST} & 0.650 &0.483 &\color{red}{0.167}\\
STARK\_RGBD \cite{kristan2021ninth} &LT &0.716&0.552 &0.164\\
DAL\cite{2019DAL} &LT &0.573 &0.410 &0.163\\
SLMD \cite{kristan2021ninth}  &LT  &0.640 &0.483 &0.157\\
TALGD \cite{kristan2021ninth} &LT  &0.666 &0.544&0.122\\
%\jinyu{SiamM\_Ds} \cite{kristan2019seventh} &0.079 &0.043 & 3.6\%\\
LTDSEd \cite{kristan2019seventh} &LT  &0.446 &0.368 &0.078\\
% OTR \cite{Kart_CVPR_2019} &LT   &0.508 &0.214 &29.4\%\\
\hline
\end{tabular}
\end{table}

\begin{figure*}
	\centering
		\includegraphics[width=0.9\linewidth]{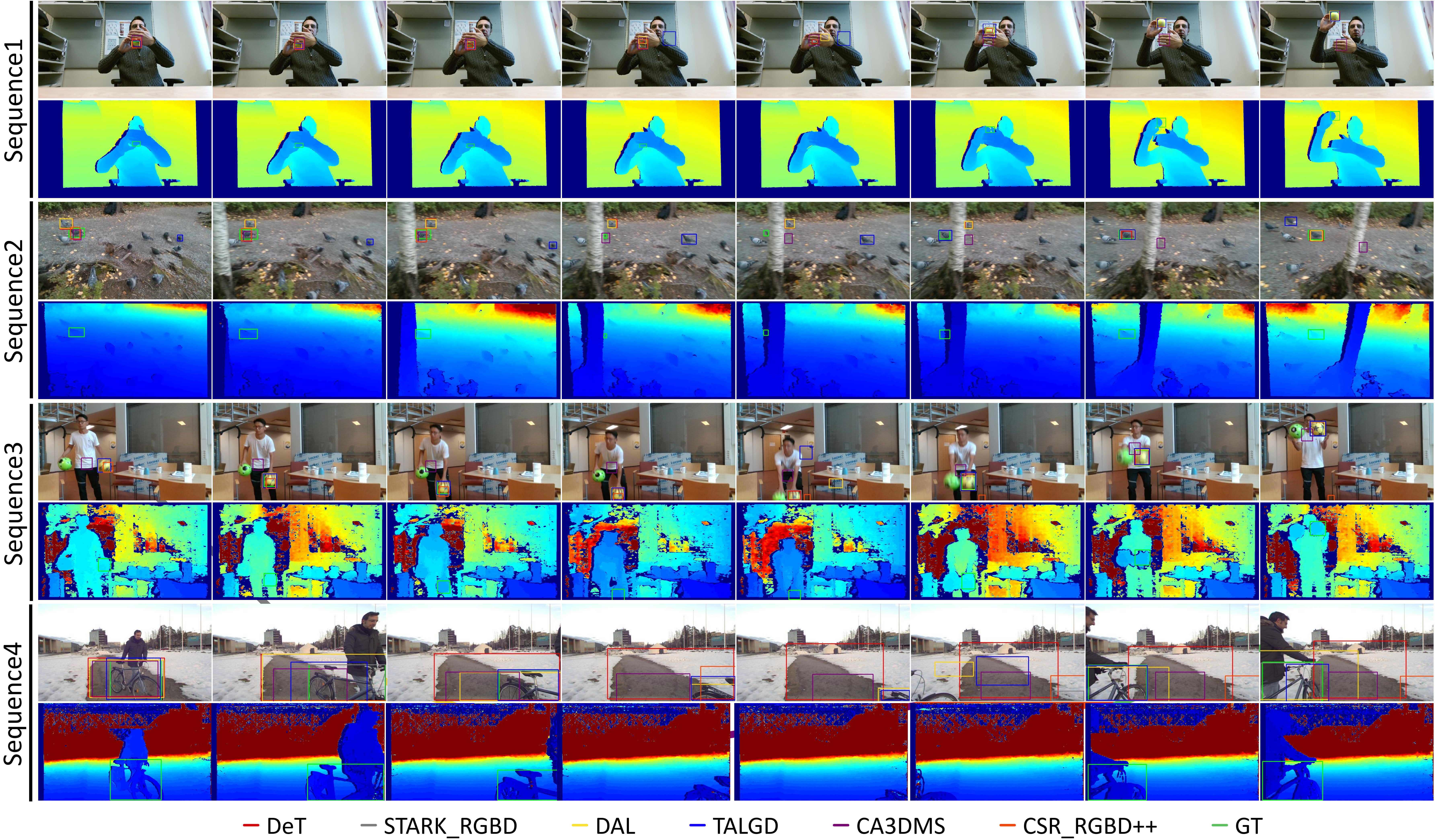}
	\caption{Visualized example sequences with long-term settings. Sequence 1-2: full occlusion; Sequence 3-4: out of view. Representative tracking results are also shown with bounding boxes in different colors.}
	\label{fig_LT}
\end{figure*}

\subsection{Long-term analysis}\label{longterm}
Long-term tracking is gaining more attention because it is much closer to practical applications than short-term tracking.
In visual object tracking, long-term tracking is defined as: the object can temporally disappear, due to occlusion or out-of-view, and re-appear in one video sequence.
In RGBD tracking, such difficulties due to the long-term settings (targets are fully occluded or out-of-view) are more common.
As shown in Table~\ref{tbl_model_statistics}, many RGBD trackers are designed for occlusion handling.
Therefore, we evaluate how state-of-the-art trackers handle the target loss and reappearance.
For evaluation, 66 sequences (98558 frames) with target loss (occlusion or out-of-view) are selected from our hybrid dataset.
The qualitative results are shown in Table~\ref{tbl_longterm}.
Generally, tracking performance degrades heavily due to target loss, especially the short-term trackers which are designed without re-detection mechanisms.
%For example, before target loss, DeT can obtain the top F-score of 0.731, but it degrades with 0.390 after target loss.
For example, DeT can get a maximum F-score of 0.731 before the target disappears but drops by 0.390 after the target disappears.
In contrast, the long-term tracker LTDSEd \cite{kristan2019seventh} degrades with only 0.078. %due to its excellent re-detection mechanism.
The stable performance of LTDSEd shows that adding depth information to the re-detection module can help relocate where objects reappear after full occlusion.
Similar to LTDSEd, TALGD \cite{kristan2021ninth} with a re-detection module can also reduce performance degradation.
% From the table, we can see that the heuristic classical long-term methods can achieve better results than the deep learning-based models but of short-term trackers.
Interestingly, traditional RGBD trackers, \textit{e.g.}, CA3DMS\cite{liu2018context}  and CSR\_RGBD++\cite{Kart_ECCVW_2018}, show heavy drops after target disappearance, although they claim that they have effective occlusion handling. CSR\_RGBD++ set some assumptions (target position and speed are basically unchanged ) in the occlusion recovery stage. CA3DMS abandoned the object detector and designed a simple strategy by comparing two depth differences to recapture the target. Consequently, the manually designed heuristic occlusion recovery modules perform poorly in the more complex and challenging datasets.
% The reasons come from their old hypothesis on tracking scenarios, which do not fit for new tracking challenges.
Therefore, equipping a re-detection module on current state-of-the-art short-term trackers can effectively improve the ability to occlusion handling. For tracking methods evaluated on long-term videos, the re-detection module is more important than using only deep features for occlusion recovery.
%It obviously indicates that the occlusion detection module using depth information in the heuristic long-term tracker models is not effective in current challenging datasets.
Some visualized examples are shown in Fig.~\ref{fig_LT}.

\subsection{Robustness evaluation}\label{robustness}
The task of object tracking can be viewed as a self-supervised learning problem due to the solely bounding box input.
Since trackers can be very sensitive to the input bounding box, tracking robustness against the input perturbations is an essential issue to judge the performance of a tracker.
However, RGBD tracking methods are compared merely using a one-pass evaluation metric for a long time.
To further investigate the performance, in this section, we propose to evaluate the robustness by comparing their performance against different perturbed inputs, such as rotation and scale, and different initialization frames.
% The initial bounding boxes for tracking are sampled spatially and temporally to evaluate the robustness and characteristics of trackers.
%The basic assumption is that a robust tracker can obtain the same performance no matter what the initial input bounding box is perturbed or set to subsequent different frames.
%Therefore, we evaluate the robustness of trackers
We choose four SOTA deep models, \textit{i.e.}, TSDM~\cite{zhao2021tsdm}, DAL~\cite{2019DAL}, DeT~\cite{yan2021depthtrack}, STARK\_RGBD~\cite{kristan2021ninth}, and two heuristic models, \textit{i.e.}, CSR\_RGBD++~\cite{kart2018depth} and DS\_KCF\_shape~\cite{2016DS}.
The detailed experimental comparison is as follows.

% \begin{table}[!t]
% \caption{Robustness evaluation with input perturbations on the hybrid benchmark.}
% \centering
% \label{tbl_dataset}
% \setlength\tabcolsep{4.5pt}
% \begin{tabular}{|c|c|cc|cc|}
% \hline
% \multicolumn{1}{|c|}{\multirow{2}{*}{Method}} & \multicolumn{1}{c|}{OPE} & \multicolumn{2}{c|}{SRE} & \multicolumn{2}{c|}{TRE} \\ \cline{2-6}
% \multicolumn{1}{|c|}{}  & F-Score & F-Score & Change & F-Score & Change\\ \hline
% TSDM\cite{zhao2021tsdm} &  0.472 &0.432 &-0.040& 0.474 & +0.002 \\
% DeT\cite{yan2021depthtrack} & 0.560 &0.512 &-0.048 & 0.578 & +0.018\\
% DAL\cite{2019DAL} &0.526 &0.484& -0.042 &0.544 &+0.018\\
% STARK\_RGBD\cite{kristan2021ninth} & 0.657 &0.606 & -0.051 &0.667 & +0.010\\
% CSR\_RGBD++\cite{kart2018depth} &0.163& 0.148 &-0.015& 0.166 & +0.003 \\
% DS\_KCF\_shape\cite{2016DS} &0.0397 &0.035 & -0.05 & 0.0396 &-0.0001\\
% \hline
% Average &0.4029 &0.3695 & -0.0334 &0.4114 & +0.0085 \\
% \hline
% \end{tabular}
% \end{table}

\begin{table}[!t]
\caption{Robustness evaluation with input perturbations on the hybrid benchmark. ``Change'' denotes the performance change compared to the original F-score.}
\centering
\label{tbl_dataset}
\setlength\tabcolsep{3pt}
\begin{tabular}{|c|c|c|c|}
\hline
\multicolumn{1}{|c|}{\multirow{2}{*}{Method}} & \multicolumn{1}{c|}{SRE} & \multicolumn{1}{c|}{TRE} & \multicolumn{1}{c|}{RTRE} \\ \cline{2-4}
\multicolumn{1}{|c|}{}  & F-Score/Change & F-Score/Change & F-Score/Change\\ \hline
DeT\cite{yan2021depthtrack} &0.512/-0.048 & 0.578/+0.018 & 0.457/-0.103\\
DAL\cite{2019DAL}  &0.484/-0.042 &0.544/+0.018 & 0.449/-0.077\\
TSDM\cite{zhao2021tsdm} &0.432/-0.040& 0.474/+0.002 & 0.323/-0.149\\
STARK\_RGBD\cite{kristan2021ninth}  &0.606/-0.051 &0.667/+0.010 & 0.580/-0.077\\
CSR\_RGBD++\cite{kart2018depth} & 0.148/-0.015& 0.166/+0.003 & 0.100/-0.063\\
DS\_KCF\_shape\cite{2016DS}  &0.035/-0.005 & 0.0396/-0.0001 & 0.0292/ -0.0105\\
\hline
Average &0.3695/-0.0334 &0.4114/+0.0085 & 0.3230/-0.0799\\
\hline
\end{tabular}
\end{table}

\subsubsection{Spatial robustness evaluation (SRE)}
%Accurate state initialization of the target object is an important factor for tracking algorithms.
%While,
In applications, the target is generally initialized manually, and thus some annotation noises will be introduced.
Thus, we usually require trackers to be stable relative to such noises.
To test the spatial robustness, we simulate the annotation error by slightly shifting or scaling the initialization state of the target.
%The performance of the algorithm in real-world application scenarios is measured by simulating the annotation error generated in the real application scenarios by slightly shifting or scaling the initialization state of the target.
We here use eight spatial shifts and four scale variations.
For the spatial shifts, we set the spatial shift as 10\% of target width, length, and corners (right, left, top, bottom shift by 10\%, and four corners shift by 10\%).
The amount for the scale variation is 10\% of the target size, and the scale ratio varies from 80\% to 120\%.
The SRE score for a tracker is the average of these 12 evaluations.
As shown in Table~\ref{tbl_dataset}, the average F-score change in SRE is -0.0334,
%Overall, deep tracking models \cite{zhao2021tsdm,2019DAL,yan2021depthtrack,kristan2021ninth} are more sensitive than the traditional ones \cite{kart2018depth,2016DS} towards input perturbations.
indicating that spatial perturbations can prevent the trackers from reliable predictions. %, and thus tracking robustness will be an important direction in the future.

\subsubsection{Temporal robustness evaluation (TRE)}
Each tracking algorithm is evaluated multiple times by setting the initial bounding box in different frames of the video sequence.
In each evaluation, the tracking algorithm starts from a specific initial frame, obtains the corresponding annotated target state, then predicts subsequent frames of the video sequence until the end of the sequence.
The average of all the test results will be taken as the TRE score of the tested algorithm.
As shown in Table~\ref{tbl_dataset}, the average F-score change is +0.0085 in the TRE, which indicates TRE has tiny effects on tracker robustness.
For SRE, the initialized target state is offset from the groundtruth bounding boxes, whereas for TRE the initialized target state is still obtained from the groundtruth bounding boxes.
%That is one of the reasons why the SRE drops after input perturbations.
Thus, the start frame changes do not affect the models very heavily.
Moreover, almost all F-scores improve slightly due to video length shortening.
%Through our observation, we find
Another reason is that researchers naturally tend to show the whole object characteristics in the very beginning since the data is collected manually, leading to the phenomenon that the tracking difficulty goes higher with video playing.
%Thus, there are only tiny fluctuations on TRE.

\subsubsection{Reverse temporal robustness evaluation (RTRE)}
From observing the RGBD video characteristics, we propose RTRE which evaluates the tracker robustness with video streams in the reverse direction.
As shown in Table~\ref{tbl_dataset}, F-scores drop severely with an average of 0.0799 due to the reverse operation of RGBD videos, indicating the test videos are more challenging in reverse order.
Overall, STARK\_RGBD\cite{kristan2021ninth} is the most robust, while DeT\cite{yan2021depthtrack} and TSDM\cite{zhao2021tsdm} drop most heavily on F-score.
%Moreover, it indicates future directions for challenging dataset design.
%the average F-score change is +0.0085 in the temporal robustness evaluation, and the average F-score change in the spatial robustness evaluation is -0.0334.
Overall, %the spatial input perturbations have a greater impact on the tracker than the temporal ones, and
deep tracking models \cite{zhao2021tsdm,2019DAL,yan2021depthtrack,kristan2021ninth} are more sensitive than the traditional ones \cite{kart2018depth,2016DS}.
%towards input perturbations.
The gap between TRE and RTRE indicates that there exists data bias in current RGBD video sequences, which can be alleviated in future dataset design.

%the collectors tend to shoot some videos that can well show the overall characteristic of the target in a relatively front part of the sequence.
% \begin{table}[!t]
% \caption{Robustness evaluation with input perturbations on the hybrid benchmark.}
% \centering
% \label{tbl_dataset}
% \begin{tabular}{|c|c|cc|cc|}
% \hline
% \multicolumn{1}{|c|}{\multirow{2}{*}{Method}} & \multicolumn{1}{c|}{OPE} & \multicolumn{2}{c|}{SRE} & \multicolumn{2}{c|}{TRE} \\ \cline{2-6}
% \multicolumn{1}{|c|}{}  & F-Score & F-Score & Change & F-Score & Change\\ \hline
% TSDM\cite{zhao2021tsdm} &  0.472 &0.432 &-0.040& 0.474 & +0.002 \\
% DeT\cite{yan2021depthtrack} & 0.560 &0.512 &-0.048 & 0.578 & +0.018\\
% DAL\cite{2019DAL} &0.526 &0.484& -0.042 &0.544 &+0.018\\
% STARK\_RGBD\cite{kristan2021ninth} & 0.657 &0.606 & -0.051 &0.667 & +0.010\\
% CSR\_RGBD++\cite{kart2018depth} &0.163& 0.148 &-0.015& 0.166 & +0.003 \\
% DS\_KCF\_shape\cite{2016DS} &0.0397 &0.035 & -0.05 & 0.0396 &-0.0001\\
% \hline
% Average &0.4029 &0.3695 & -8.30\% &0.4114 & +2.10\% \\
% \hline
% \end{tabular}
% \end{table}

\section{Discussions}\label{discussion}
%In this section, we discuss several potential research directions for RGBD tracking.

%can be used to be addressed to further narrow down the existing gap.
%\subsection{Main challenges}
\subsection{Dataset construction}

\subsubsection{High-quality dataset}
Based on the experiments in Sec.~\ref{quality}, we conclude that low depth quality has side effects on tracking performance.
To develop RGBD object tracking, collecting high-quality data resources is very important.
% is of vital importance to the RGBD tracking and independent depth tracking.
Due to the limitations of depth sensors, the collection range of depth maps is lower than 10m, in which we can only get precise depth maps between 0.5m-6m.
Thus, the pre-processing of low-quality depth videos is also an important step.
In addition, the synchronization and registration noise also damages tracking performance as well.
Thus, both reliable depth information and correctly synchronized RGBD data are necessary for trackers to effectively combine the multi-modal information.

% \subsubsection{Large-scale dataset}
% The size of existing RGBD datasets for tracking is quite limited compared to RGB ones.
% Thus, it is necessary to develop new large-scale datasets for data-hungry RGBD tracking models.

\subsubsection{Domain-specific dataset}
Due to different application scenarios, there are datasets specifically designed for different applications, \textit{e.g.}, UAV tracking, small-sized object tracking, \textit{etc}.
However, there are no such datasets for RGBD tracking.
As RGBD tracking has shown potential on a wide range of applications, collecting such domain-specific datasets may benefit the application of it on specific scenarios, \textit{e.g.}, nighttime scenarios and UAV tracking scenarios.

\subsubsection{Data annotation}
Advanced RGB-based tracking benchmarks are transferred to more accurate and fine-grained annotation with mask annotations, while RGBD benchmark datasets remain on axis-aligned bounding boxes.
The bounding boxes prevent RGBD trackers from better describing the object, especially the irregularly shaped and deformable objects.
A potential direction of RGBD tracking is to develop pixel-level RGBD tracking algorithms, which indeed need more precise data annotations.
\subsubsection{Challenging videos}
Compared with the RGB dataset, which can be randomly intercepted from the Internet, RGBD datasets are mostly manually captured by the researchers. % is more scarce.
%From our observations of the datasets, collectors will subconsciously select some frames that better represent the overall characteristics of the target as the initial frame.
As we observed in Sec.~\ref{dataset_metric} and Sec.~\ref{robustness}, target locations are often around the center and target states are mostly stationary or moving slowly in video beginning, which results in unfair comparison.
Thus, randomly cropped RGBD videos will alleviate the data bias between video beginning and ending parts, so they are more challenging for trackers.
Also, RTRE is effective to evaluate the tracker's robustness and data bias.%to keep an accurate and robust prediction.
%For the challenge of the dataset, we think it is necessary to choose some targets with more random positions and natural motion states as the target of the initial frame.

\subsection{Evaluation protocols}
A wealth of performance measures have been proposed for RGBD tracker evaluation, but they are proposed accompanied by different datasets with different characteristics.
For example, STC employs evaluation metrics from OTB and VOT, which are both determined to short-term tracking and not suitable for long-term tracking.
CDTB's and DepthTrack's evaluation systems are modified from long-term tracking, requiring the trackers to report both bounding box and confidence score which early trackers cannot.
Thus, a unified evaluation protocol is necessary which can clearly reflect different aspects of tracking.
%Apart from merely ranking, we also need to determine cases when two or more trackers are performing equally well.
Therefore, RGBD tracking requires measures to allow easy interpretation and tracker comparison with a well-defined tracker equivalence.

\subsection{Model design}
%Based on the review of RGBD tracking algorithms in Sec.~\ref{model}, we discuss potential directions for RGBD tracker design.

\subsubsection{New paradigm}
Novel paradigms have been introduced into many vision tasks, including RGB-based object tracking.
For example, the modified version of the transformer-based RGB tracking model STARK \cite{yan2021learning} shows extraordinary performance on RGBD tracking, which indicates that the novel paradigm is of importance in multi-modal tracking.
Therefore, importing novel paradigms from related areas is of potential. %for high-performance RGBD tracking.
%At the same time, efficient RGBD models with high-speed-performance trade-offs are also welcome.

\subsubsection{Cross-modal fusion}
It is important to effectively fuse RGB and depth information in RGBD tracking.
Existing tracking models mostly employ heuristic fusion strategies, \textit{e.g.}, the weighted sum between two modalities, modality selection, and joint distribution.
Effective trainable deep features and fusion modules are required.

\subsubsection{Model efficiency}
As object tracking is a real-time application, tracking speed is an important measure, which is ignored in RGBD tracking for a long time.
We notice that most RGBD trackers' speed is out of real-time requirement.
When VOT participants obtain good performance on the hybrid datasets and several sub-tests, their speed is relatively low.
Getting a good trade-off between performance and speed is still challenging to RGBD model design.
Lightweight architectures are required for real-world applications.

\subsubsection{Model robustness}
From experiments, we observed that, although deep models show high tracking performance, they are not robust to both spatial and temporal perturbations, which may prevent the models to be used in real-life applications.
Also, adversarial attacks and the defense of RGBD tracking algorithms are required to be studied.
We hope our observations will shed light on the model robustness research.

\section{Conclusions}\label{conclusion}
In this paper, we present a comprehensive review of RGBD tracking.
To this end, we first provide taxonomies for categorizing RGBD tracking models, including depth usage, heuristic/deep models, and RGBD fusion.
Then, we cover the popular datasets in RGBD tracking and review the related evaluation metrics.
Next, by offering a hybrid dataset containing the public RGBD tracking sequences, we can evaluate 18 RGBD trackers for performance benchmarking.
Most importantly, we conduct extensive experiments and provide in-depth analysis on the representative trackers from overall performance, depth data quality, long-term settings, attribute-based tests, and robustness evaluation.
Finally, discussions on dataset construction, evaluation protocols, and model design are given for further research.

% \section{References Section}
% You can use a bibliography generated by BibTeX as a .bbl file.
%  BibTeX documentation can be easily obtained at:
%  http://mirror.ctan.org/biblio/bibtex/contrib/doc/
%  The IEEEtran BibTeX style support page is:
%  http://www.michaelshell.org/tex/ieeetran/bibtex/

\bibliographystyle{IEEEtran}

\bibliography{cas-refs}

\end{document}